\documentclass[10pt,twocolumn,letterpaper]{article}

\usepackage{cvpr}
\usepackage[utf8]{inputenc}
\usepackage{times}
\usepackage{epsfig}
\usepackage{graphicx}
\usepackage{amsmath}
\usepackage{amssymb}
\usepackage{booktabs}
\usepackage{adjustbox}
\usepackage{scalefnt}
\usepackage{fancyhdr}
\usepackage{subcaption}
\usepackage{color,soul}

\usepackage[pagebackref=true,breaklinks=true,letterpaper=true,colorlinks,bookmarks=false]{hyperref}

\cvprfinalcopy 

\ifcvprfinal\fi
\begin{document}

\title{3D human pose estimation in video with temporal convolutions and \\ semi-supervised training}

\author{Dario Pavllo\thanks{Work done while at Facebook AI Research.}\\
ETH Zürich\\
\and
Christoph Feichtenhofer\\
Facebook AI Research\\
\and
David Grangier\footnotemark[1]\\
Google Brain\\
\and
Michael Auli\\
Facebook AI Research\\
}

\newcommand{\macomment}[1]{\textcolor{blue}{{*MA*: #1}}}
\newcommand{\mamark}[1]{\textcolor{blue}{{#1}}}
\newcommand{\dpcomment}[1]{\textcolor{red}{{*DP*: #1}}}
\newcommand{\dpmark}[1]{\textcolor{red}{{#1}}}
\newcommand{\cfcomment}[1]{\textcolor{blue}{{*CF*: #1}}}
\newcommand{\cfmark}[1]{\textcolor{blue}{{#1}}}

\maketitle
\thispagestyle{empty}

\begin{abstract}
    In this work, we demonstrate that 3D poses in video can be effectively estimated with a fully convolutional model based on dilated temporal convolutions over 2D keypoints.
    We also introduce back-projection, a simple and effective semi-supervised training method that leverages unlabeled video data.
    We start with predicted 2D keypoints for unlabeled video, then estimate 3D poses and finally back-project to the input 2D keypoints.
    In the supervised setting, our fully-convolutional model outperforms the previous best result from the literature by 6 mm mean per-joint position error on Human3.6M, corresponding to an error reduction of 11\%, and the model also shows significant improvements on HumanEva\nobreakdash-I.
    Moreover, experiments with back-projection show that it comfortably outperforms previous state-of-the-art results in semi-supervised settings where labeled data is scarce. Code and models are available at \url{https://github.com/facebookresearch/VideoPose3D}
\end{abstract}

\section{Introduction}
Our work focuses on 3D human pose estimation in video. 
We build on the approach of
state-of-the-art methods which formulate the problem as 2D
keypoint detection followed by 3D pose estimation \cite{pavlakos:coarse:2017,tekin:learning:2017,martinez:simple:2017,sun:compositional:2017,fang:learning:2018,pavlakos:ordinal:2018,yang:3d:2018,luvizon:2d:2018}.
While
splitting up the problem arguably reduces the difficulty of
the task, it is inherently ambiguous as multiple 3D poses
can map to the same 2D keypoints.
Previous work tackled this ambiguity by modeling temporal information with recurrent neural networks~\cite{hossain:exploiting:2018,lee:propagating:2018}.
On the other hand, convolutional networks have been very successful in modeling temporal information in tasks that were traditionally tackled with RNNs, such as neural machine translation \cite{gehring:convs2s:2017}, language modeling \cite{dauphin:gatedlm:2017}, speech generation \cite{oord:wavenet:2016}, and speech recognition \cite{collobert:wav2letter:2016}.
Convolutional models enable parallel processing of multiple frames which is not possible with recurrent networks.

\begin{figure}[t]
	\centering
	\vspace{10pt}
	\includegraphics[width=1.0\linewidth]{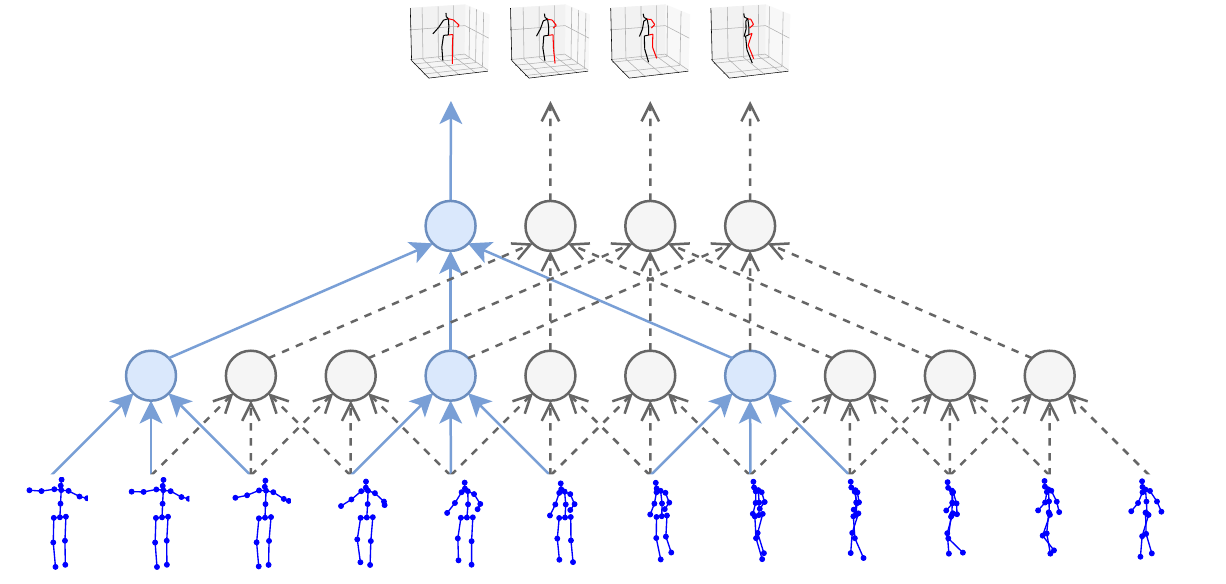}
	\caption{Our temporal convolutional model takes 2D keypoint sequences (bottom) as input and generates 3D pose estimates as output (top). We employ dilated temporal convolutions to capture long-term information. }
    \label{fig:convolutions}
\end{figure}

In this paper, we present a fully convolutional architecture that performs temporal convolutions over 2D keypoints for accurate 3D pose prediction in video (see Figure \ref{fig:convolutions}).
Our approach is compatible with any 2D keypoint detector and can effectively handle large contexts via dilated convolutions. 
Compared to approaches relying on RNNs \cite{hossain:exploiting:2018,lee:propagating:2018}, it provides higher accuracy, simplicity, as well as efficiency, both in terms of computational complexity as well as the number of parameters (\textsection\ref{sec:approach_conv}).

Equipped with a highly accurate and efficient architecture, we turn to settings where labeled training data is scarce and introduce a new scheme to leverage unlabeled video data for \emph{semi-supervised training}.
Low resource settings are particularly challenging for neural network models which require large amounts of labeled training data and collecting labels for 3D human pose estimation requires an expensive motion capture setup as well as lengthy recording sessions.
Our method is inspired by cycle consistency in unsupervised machine translation, where round-trip translation into an intermediate language and back into the original language should be close to the identity function~\cite{sennrich2016bpe,lample2017unsupervised,edunov2018bt}.
Specifically, we predict 2D keypoints for an unlabeled video with an off the shelf 2D keypoint detector, predict 3D poses, and then map these back to 2D space (\textsection\ref{sec:semi}).

In summary, this paper  provides two main contributions.
First, we present a simple and efficient approach for 3D human pose estimation in video based on dilated temporal convolutions on 2D keypoint trajectories. We show that our model is more efficient than RNN-based models at the same level of accuracy, both in terms of computational complexity and the number of model parameters.

Second, we introduce a semi-supervised approach which exploits unlabeled video, and is effective when labeled data is scarce. Compared to previous semi-supervised approaches, we only require camera intrinsic parameters rather than ground-truth 2D annotations or multi-view imagery with extrinsic camera parameters.

In comparison to the state of the art our approach outperforms the previously best performing methods in both supervised and semi-supervised settings. Our supervised model performs better than other models even if these exploit extra labeled data for training.



    
    
    
    

\section{Related work}
\label{sec:pose_related}
Before the success of deep learning, most approaches to 3D pose estimation were based on feature engineering and assumptions about skeletons and joint mobility \cite{sminchisescu:3d:2008, ramakrishna:reconstructing:2012, ionescu:human36:2014, ionescu:iterated:2014}.
The first neural methods with convolutional neural networks (CNN) focused on end-to-end reconstruction \cite{li:3d:2014, tekin:direct:2016, tekin:structured:2016, pavlakos:coarse:2017} by directly estimating 3D poses from RGB images without intermediate supervision.

\noindent\textbf{Two-step pose estimation.}
A new family of 3D pose estimators builds on top of 2D pose estimators by first predicting 2D joint positions in image space (\emph{keypoints}) which are subsequently lifted to 3D~\cite{jiang:3d:2010, martinez:simple:2017, pavlakos:coarse:2017, tekin:learning:2017, chen:3d:2017, hossain:exploiting:2018}.
These approaches outperform the end-to-end counterparts, since they benefit from intermediate supervision.
We follow this approach.
Recent work shows that predicting 3D poses is relatively straightforward given ground-truth 2D keypoints, and that the difficulty lies in predicting accurate 2D poses~\cite{martinez:simple:2017}.
Early approaches~\cite{jiang:3d:2010, chen:3d:2017} simply perform a k-nearest neighbour search for a predicted set of 2D keypoints over a large set of 2D keypoints for which the 3D pose is available and then simply output the corresponding 3D pose.
Some approaches leverage both image features and 2D ground-truth poses~\cite{park:3d:2016, pavlakos:coarse:2017, tekin:learning:2017}.
Alternatively, the 3D pose can be predicted from a given set of 2D keypoints by simply predicting their depth~\cite{zhou:towards:2017}. Some works enforce priors about bone lengths and projection consistency with the 2D ground truth \cite{brau:3d:2016}.

\noindent\textbf{Video pose estimation.}
Most previous work operates in a single-frame setting but recently there have been efforts in exploiting temporal information from video to produce more robust predictions and to be less sensitive to noise. 
\cite{tekin:direct:2016} infer 3D poses from the HoG features (histograms of oriented gradients) of spatio-temporal volumes. 
LSTMs have been used to refine 3D poses predicted from single images~\cite{lin:recurrent:2017, katircioglu:learning:2018}.
The most successful approaches, however, learn from \emph{2D keypoint trajectories}. 
Our work falls under this category.

Recently, LSTM sequence-to-sequence learning models have been proposed, which encode a sequence of 2D poses from a video into a fixed-size vector that is then decoded into a sequence of 3D poses~\cite{hossain:exploiting:2018}.
However, both the input and output sequences have the same length and a deterministic transformation of 2D poses is a much more natural choice.
Our experiments with \emph{seq2seq} models showed that output poses tend to drift over lengthy sequences. 
\cite{hossain:exploiting:2018} tackles this problem by re-initializing the encoder every 5 frames, at the expense of temporal consistency. 
There has also been work on RNN approaches which consider priors on body part connectivity~\cite{lee:propagating:2018}.

\noindent\textbf{Semi-supervised training.}
There has been work on multitask networks~\cite{caruana:multitask:1997} for joint 2D and 3D pose estimation \cite{mehta:vnect:2017, luvizon:2d:2018} as well as action recognition~\cite{luvizon:2d:2018}. 
Some works transfer the features learned for 2D pose estimation to the 3D task \cite{mehta:monocular:2017}.
Unlabeled multi-view recordings have been used for pre-training representations for 3D pose estimation~\cite{rhodin:unsupervised:2018}, but these recordings are not readily available in unsupervised settings.
Generative adversarial networks (GAN) can discriminate realistic poses from unrealistic ones in a second dataset where only 2D annotations are available~\cite{yang:3d:2018}, thus providing a useful form of regularization. \cite{tung:adversarial:2017} use GANs to learn from unpaired 2D/3D datasets and include a 2D projection consistency term. Similarly, \cite{drover:2dprojections:2018} discriminate generated 3D poses after randomly projecting them to 2D.
\cite{pavlakos:ordinal:2018} propose a weakly-supervised approach based on ordinal depth annotations which leverages a 2D pose dataset augmented with depth comparisons, e.g. ``the left leg is behind the right leg''. 

\noindent\textbf{3D shape recovery.}
While this paper and the discussed related work focus on reconstructing accurate 3D poses, a parallel line of research aims at recovering full 3D shapes of people from images \cite{bogo:keepitsmpl:2016, kanazawa:shapepose:2018}. These approaches are typically based on parameterized 3D meshes and give less importance to pose accuracy.

\begin{figure*}
	\centering
	\includegraphics[width=\textwidth]{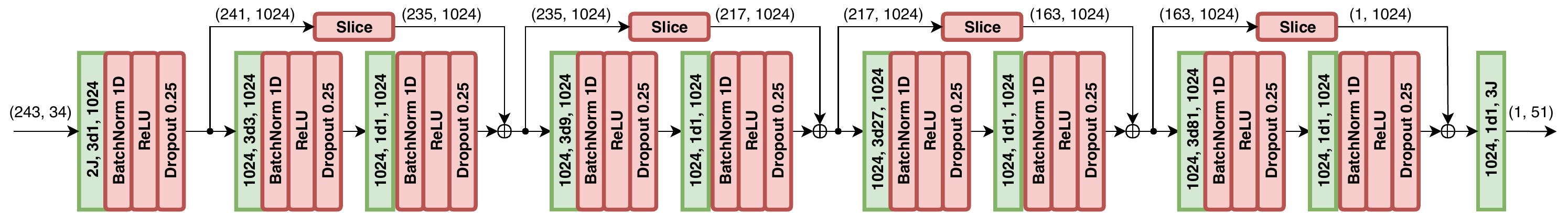}
	\caption{An instantiation of our fully-convolutional 3D pose estimation architecture. The input consists of 2D keypoints for a recpetive field of 243 frames ($B = 4$ blocks) with $J = 17$ joints.
	Convolutional layers are in green where \texttt{2J, 3d1, 1024} denotes $2 \cdot J$ input channels, kernels of size 3 with dilation 1, and 1024 output channels. 
	We also show tensor sizes in parentheses for a sample 1-frame prediction, where \texttt{(243, 34)} denotes 243 frames and 34 channels. 
	Due to valid convolutions, we slice the residuals (left and right, symmetrically) to match the shape of subsequent tensors. }
    \label{fig:3d_model}
\end{figure*}

\noindent\textbf{Our work.}
Compared to \cite{pavlakos:coarse:2017, pavlakos:ordinal:2018}, we do not use heatmaps and instead describe poses with detected keypoint coordinates. This allows the use of efficient 1D convolution over coordinate time series, instead of 2D convolutions over individual heatmaps (or 3D convolutions over heatmap sequences). Our approach also makes computational complexity independent of keypoint spatial resolution.
Our models can reach high accuracy with fewer parameters and allow for faster training and inference. 
Compared to the single-frame baseline proposed by \cite{martinez:simple:2017} and the LSTM model by \cite{hossain:exploiting:2018}, we exploit temporal information by performing 1D convolutions over the time dimension, and we propose several optimizations that result in lower reconstruction error. Unlike \cite{hossain:exploiting:2018}, we learn a deterministic mapping instead of a \emph{seq2seq} model. 
Finally, contrary to most of the two-step models mentioned in this section (which use the popular \emph{stacked hourglass network} \cite{newell:stacked:2016} for 2D keypoint detection), we show that Mask R-CNN \cite{he:mask:2017} and cascaded pyramid network (CPN) \cite{chen:cascaded:2018} detections are more robust for 3D human pose estimation.

\section{Temporal dilated convolutional model} 
\label{sec:approach_conv}

Our model is a fully convolutional architecture with residual connections that takes a sequence of 2D poses as input and transforms them through temporal convolutions.
Convolutional models enable parallelization over both the batch and the time dimension while RNNs cannot be parallelized over time.
In convolutional models, the path of the gradient between output and input has a fixed length regardless of the sequence length, which mitigates vanishing and exploding gradients which affect RNNs. 
A convolutional architecture also offers precise control over the temporal receptive field, which we found beneficial to model temporal dependencies for the task of 3D pose estimation. 
Moreover, we employ \emph{dilated convolutions}~\cite{holschneider:wavelet:1989} to model long-term dependencies while at the same time maintaining efficiency. 
Architectures with dilated convolutions have been successful for audio generation \cite{oord:wavenet:2016}, semantic segmentation \cite{yu:multi:2015} and machine translation~\cite{kalchbrenner:bytenet:2016}. 

The input layer takes the concatenated $(x,y)$ coordinates of the $J$ joints for each frame and applies a temporal convolution with kernel size $W$ and $C$ output channels. 
This is followed by $B$ ResNet-style blocks which are surrounded by a skip-connection \cite{he:deep:2016}.
Each block first performs a 1D convolution with kernel size $W$ and dilation factor $D = W^B$, followed by a convolution with kernel size 1.
Convolutions (except the very last layer) are followed by batch normalization \cite{ioffe:batch:2015}, rectified linear units~\cite{nair2010rectified}, and dropout~\cite{srivastava:dropout:2014}.
Each block increases the receptive field exponentially by a factor of $W$, while the number of parameters increases only linearly. 
The filter hyperparameters, $W$ and $D$, are set so that the receptive field for any output frame forms a tree that covers all input frames (see \textsection\ref{fig:convolutions}). 
Finally, the last layer outputs a prediction of the 3D poses for all frames in the input sequence using both past and future data to exploit temporal information.
To evaluate real-time scenarios, we also experiment with \emph{causal convolutions}, \ie convolutions that only have access to past frames. Appendix~\ref{app:flow}
illustrates dilated convolutions and causal convolutions.

Convolutional image models typically apply zero-padding to obtain as many outputs as inputs. Early experiments however showed better results when performing only unpadded convolutions while padding the input sequence with replica of the boundary frames to the left and the right (see Appendix~\ref{app:batching}, Figure~\ref{fig:batching} for an illustration). 

Figure~\ref{fig:3d_model} shows an instantiation of our architecture for a receptive field size of \emph{243 frames} with $B=4$ blocks.
For convolutional layers, we set $W=3$ with $C=1024$ output channels and we use a dropout rate $p=0.25$.

\section{Semi-supervised approach}
\label{sec:semi}

We introduce a semi-supervised training method to improve accuracy in settings where the availability of labeled 3D ground-truth pose data is limited.
We leverage \emph{unlabeled} video in combination with an off the shelf 2D keypoint detector to extend the supervised loss function with a \emph{back-projection} loss term. 
We solve an auto-encoding problem on unlabeled data: the encoder (pose estimator) performs 3D pose estimation from 2D joint coordinates and the decoder (projection layer) projects the 3D pose back to 2D joint coordinates. Training penalizes when the 2D joint coordinates from the decoder are far from the original input.

\begin{figure}
	\centering
    \includegraphics[width=\linewidth]{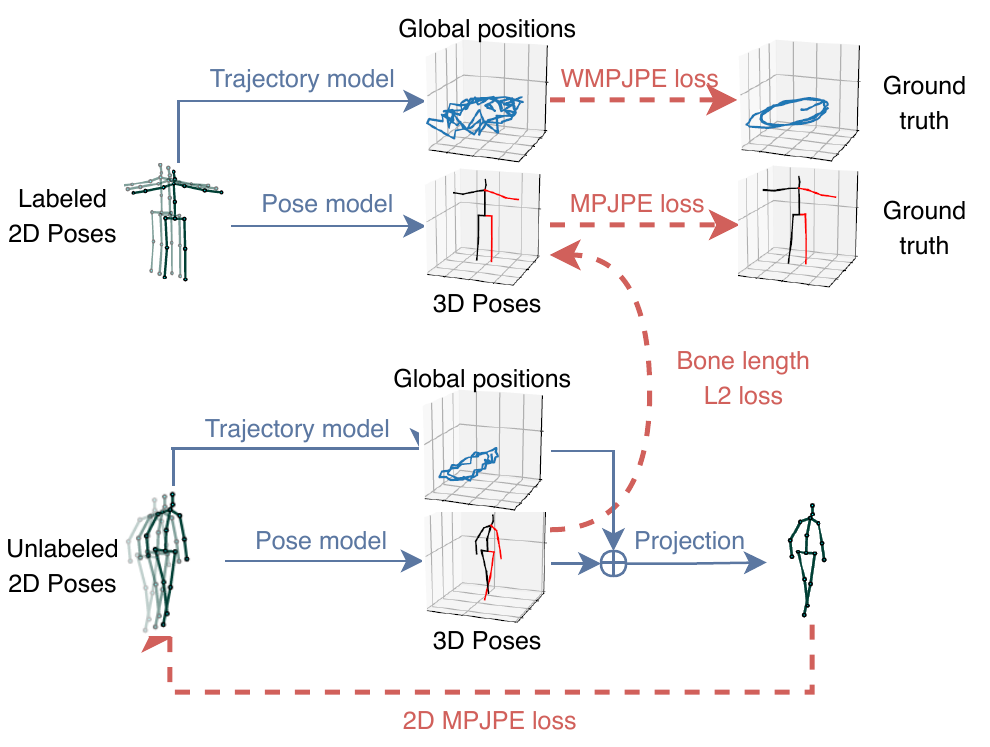}
	\caption{Semi-supervised training with a 3D pose model that takes a sequence of possibly predicted 2D poses as input. 
	We regress the 3D trajectory of the person and add a soft-constraint to match the mean bone lengths of the unlabeled predictions to the labeled ones.
	Everything is trained jointly. WMPJPE stands for ``Weighted MPJPE''.}
	\label{fig:semi_pipeline}
\end{figure}

Figure~\ref{fig:semi_pipeline} represents our method which combines our supervised component with our unsupervised component which acts as a regularizer.
The two objectives are optimized jointly, with the labeled data occupying the first half of a batch, and the unlabeled data occupying the second half. 
For the labeled data we use the ground truth 3D poses as target and train a supervised loss.
The unlabeled data is used to implement an autoencoder loss where the predicted 3D poses are projected back to 2D and then checked for consistency with the input. 

\noindent\textbf{Trajectory model.}
Due to the perspective projection, the 2D pose on the screen depends both on the trajectory (i.e. the global position of the human root joint) and the 3D pose (the position of all joints with respect to the root joint). Without the global position, the subject would always be reprojected at the center of the screen with a fixed scale.
We therefore also regress the 3D trajectory of the person, so that the back-projection to 2D can be performed correctly. To this end, we optimize a second network which regresses the global trajectory \emph{in camera space}. 
The latter is added to the pose before projecting it back to 2D. 
The two networks have the same architecture but do not share any weights as we observed that they affect each other negatively when trained in a multi-task fashion. 
As it becomes increasingly difficult to regress a precise trajectory if the subject is further away from the camera, we optimize a weighted mean per-joint position error (WMPJPE) loss function for the trajectory:
\begin{align}
    E = \frac{1}{\textbf{y}_z} \lVert f(\textbf{x}) - \textbf{y} \rVert
\end{align}
that is, we weight each sample using the inverse of the ground-truth depth ($\textbf{y}_z$) in camera space. 
Regressing a precise trajectory for far subjects is also unnecessary for our purposes, since the corresponding 2D keypoints tend to concentrate around a small area.

\noindent\textbf{Bone length L2 loss}.
We would like to incentivize the prediction of plausible 3D poses instead of just copying the input.
To do so, we found it effective to add a soft constraint to approximately match the mean bone lengths of the subjects in the unlabeled batch to the subjects of the labeled batch (``Bone length L2 loss'' in Figure~\ref{fig:semi_pipeline}). 
This term plays an important role in self-supervision, as we show in \textsection\ref{sec:semi_eval}.

\noindent\textbf{Discussion}.
Our method only requires the camera \emph{intrinsic} parameters, which are often available for commercial cameras.\footnote{Even low-end devices typically embed this information in the EXIF metadata of images or videos.}
The approach is not tied to any specific network architecture and can be applied to any 3D pose detector which takes 2D keypoints as inputs. 
In our experiments we use the architecture described in \textsection\ref{sec:approach_conv} to map 2D poses to 3D.
To project 3D poses to 2D, we use a simple projection layer which considers linear parameters (focal length, principal point) as well as non-linear lens distortion coefficients (tangential and radial).
We found the lens distortions of the cameras used in Human3.6M have negligible impact on the pose estimation metric, but we include these terms nonetheless because they always provide a more accurate modeling of the real camera projection.

\section{Experimental setup}
\label{sec:setup}

\subsection{Datasets and Evaluation}
We evaluate on two motion capture datasets, Human3.6M \cite{ionescu:human36:2014, ionescu:iccv:2011} and HumanEva-I \cite{sigal:humaneva:2010}.
{Human3.6M} contains 3.6 million video frames for 11 subjects, of which seven are annotated with 3D poses. 
Each subject performs 15 actions that are recorded using four synchronized cameras at 50 Hz. 
Following previous work \cite{pavlakos:coarse:2017, tekin:learning:2017, martinez:simple:2017, sun:compositional:2017, fang:learning:2018, pavlakos:ordinal:2018, yang:3d:2018, luvizon:2d:2018}, we adopt a 17-joint skeleton, train on five subjects (S1, S5, S6, S7, S8), and test on two subjects (S9 and S11).
We train a single model for all actions. 

{HumanEva-I} is a much smaller dataset, with three subjects recorded from three camera views at 60 Hz. 
Following~\cite{martinez:simple:2017, hossain:exploiting:2018}, we evaluate on three actions (Walk, Jog, Box) by training a different model for each action (\emph{single action} -- SA).
We also report results when training one model for all actions (\emph{multi action} -- MA), as in~\cite{pavlakos:coarse:2017, lee:propagating:2018}.
We adopt a 15-joint skeleton and use the provided train/test split.

In our experiments, we consider three evaluation protocols:
\textbf{Protocol 1} is the mean per-joint position error (MPJPE) in millimeters which is the mean Euclidean distance between predicted joint positions and ground-truth joint positions and follows~\cite{li:maximum:2015, tekin:direct:2016, zhou:sparseness:2016, martinez:simple:2017, pavlakos:coarse:2017}.
\textbf{Protocol 2} reports the error after alignment with the ground truth in translation, rotation, and scale (P-MPJPE)~\cite{martinez:simple:2017, sun:compositional:2017, fang:learning:2018, pavlakos:ordinal:2018, yang:3d:2018, hossain:exploiting:2018}. 
\textbf{Protocol 3} aligns predicted poses with the ground-truth only in scale (N-MPJPE) following~\cite{rhodin:unsupervised:2018} for semi-supervised experiments.

\subsection{Implementation details for 2D pose estimation}
\label{sec:impl_keypoint}

Most previous work~\cite{martinez:simple:2017, zhou:towards:2017, tekin:learning:2017} extracts the subject from ground-truth bounding boxes and then applies the stacked hourglass detector to predict the 2D keypoint locations within the ground-truth bounding box~\cite{newell:stacked:2016}. 
Our approach (\textsection\ref{sec:approach_conv} and \textsection\ref{sec:semi}) does not depend on any particular 2D keypoint detector. 
We therefore investigate several 2D detectors that do not rely on ground-truth boxes which enables the use of our setup in the wild.
In addition to the \emph{stacked hourglass detector}, we investigate \emph{Mask R-CNN} \cite{he:mask:2017} with a ResNet-101-FPN~\cite{lin:feature:2017} backbone, using its reference implementation in \emph{Detectron}, as well as \emph{cascaded pyramid network} (CPN) \cite{chen:cascaded:2018} which represents an extension of FPN. The CPN implementation requires bounding boxes to be provided externally (we use Mask R-CNN boxes for this case). 

For both Mask R-CNN and CPN, we start with pretrained models on COCO \cite{lin:coco:2014} and fine-tune the detectors on 2D projections of Human3.6M, since the keypoints in COCO differ from Human3.6M \cite{ionescu:human36:2014}.
In our ablations, we also experiment with directly applying our 3D pose estimator to pretrained 2D COCO keypoints for estimating the 3D joints of Human3.6M.

For Mask R-CNN, we adopt a ResNet-101 backbone trained with the ``stretched 1x'' schedule \cite{he:mask:2017}.\footnote{ \url{https://github.com/facebookresearch/Detectron/blob/master/configs/12_2017_baselines/e2e_keypoint_rcnn_R-101-FPN_s1x.yaml}}
When fine-tuning the model on Human3.6M, we reinitialize the last layer of the keypoint network, as well as the {deconv} layers that regress the heatmaps to learn a new set of keypoints. 
We train on 4 GPUs with a step-wise decaying learning rate: 1e-3 for 60k iterations, then 1e-4 for 10k iterations, and 1e-5 for 10k iterations. 
At inference, we apply a softmax over the the heatmaps and extract the expected value of the resulting 2D distribution (\emph{soft-argmax}).
This results in smoother and more precise predictions than \emph{hard-argmax} \cite{luvizon:2d:2018}. 

For CPN, we use a ResNet-50 backbone with a 384$\times$288 resolution. 
To fine-tune, we re-initialize the final layers of both \emph{GlobalNet} and \emph{RefineNet} (convolution weights and batch normalization statistics). 
Next, we train on one GPU with batches of 32 images and with a step-wise decaying learning rate: 5e-5 (1/10th of the initial value) for 6k iterations, then 5e-6 for 4k iterations, and finally 5e-7 for 2k iterations. 
We keep batch normalization enabled while fine-tuning. 
We train with ground-truth bounding boxes and test using the bounding boxes predicted by the fine-tuned Mask R-CNN model.

\subsection{Implementation details for 3D pose estimation}
For consistency with other work~\cite{martinez:simple:2017, li:maximum:2015, tekin:direct:2016, zhou:sparseness:2016, martinez:simple:2017, pavlakos:coarse:2017}, we train and evaluate on 3D poses in \emph{camera space} by rotating and translating the ground-truth poses according to the camera transformation, and not using the global trajectory (except for the semi-supervised setting, \textsection\ref{sec:semi}).

As optimizer we use Amsgrad \cite{reddi:amsgrad:2018} and train for 80 epochs.
For Human3.6M, we adopt an exponentially decaying learning rate schedule, starting from $\eta = 0.001$ with a shrink factor $\alpha = 0.95$ applied each epoch. 

All temporal models, \ie models with receptive fields larger than one, are sensitive to the correlation of samples in pose sequences (\emph{cf.~}\textsection\ref{sec:approach_conv}). 
This results in biased statistics for batch normalization which assumes independent samples \cite{ioffe:batch:2015}.
In preliminary experiments, we found that predicting a large number of adjacent frames during training yields results that are worse than a model exploiting no temporal information (which has well-randomized samples in the batch). 
We reduce correlation in the training samples by choosing training clips from different video segments. 
The clip set size is set to the width of the receptive field of our architecture so that the model predicts a single 3D pose per training clip.
This is important for generalization and we analyze it in detail in Appendix~\ref{app:batching}. 

We can greatly optimize this single frame setting by replacing dilated convolutions with strided convolutions where the stride is set to be the dilation factor (see Appendix~\ref{app:optimize}).
This avoids computing states that are never used and we apply this optimization only during training.
At inference, we can process entire sequences and reuse intermediate states of other 3D frames for faster inference. 
This is possible because our model does not use any form of pooling over the time dimension.
To avoid losing frames to valid convolutions, we pad by replication, but only at the input boundaries of a sequence (Appendix~\ref{app:batching}, Figure~\ref{fig:batching} shows an illustration). 

We observed that the default hyperparameters of batch normalization lead to large fluctuations of the test error ($\pm$ 1 mm) as well as to fluctuations in the running estimates for inference. 
To achieve more stable running statistics, we use a schedule for the batch-normalization momentum $\beta$: we start from $\beta = 0.1$, and decay it exponentially so that it reaches $\beta = 0.001$ in the last epoch.

Finally, we perform horizontal flip augmentation at train and test time. 
We show the effect of this in Appendix~\ref{app:data_conv}.

For HumanEva, we use $N = 128$, $\alpha = 0.996$, and train for 1000 epochs using a receptive field of 27 frames. 
Some frames in HumanEva are corrupted by sensor dropout and we split the corrupted videos into valid contiguous chunks and treat them as independent videos.

\begin{table*}[t]
    \begin{subtable}{\linewidth}
        \centering
		\small
    	\tabcolsep=1mm
    	\resizebox{\textwidth}{!}{
    		\begin{tabular}{@{}l|rrrrrrrrrrrrrrr|r@{}}
    		& Dir. & Disc. & Eat & Greet & Phone & Photo & Pose & Purch. & Sit & SitD. & Smoke & Wait & WalkD. & Walk & WalkT. & Avg\\
    		\midrule
    		Pavlakos \etal \cite{pavlakos:coarse:2017} CVPR'17 $(\ast)$ & 67.4 & 71.9 & 66.7 & 69.1 & 72.0 & 77.0 & 65.0 & 68.3 & 83.7 & 96.5 & 71.7 & 65.8 & 74.9 & 59.1 & 63.2 & 71.9 \\
    		Tekin \etal \cite{tekin:learning:2017} ICCV'17 & 54.2 & 61.4 & 60.2 & 61.2 & 79.4 & 78.3 & 63.1 & 81.6 & 70.1 & 107.3 & 69.3 & 70.3 & 74.3 & 51.8 & 63.2 & 69.7 \\
    		Martinez \etal \cite{martinez:simple:2017} ICCV'17 $(\ast)$ & 51.8 & 56.2 & 58.1 & 59.0 & 69.5 & 78.4 & 55.2 & 58.1 & 74.0 & 94.6 & 62.3 & 59.1 & 65.1 & 49.5 & 52.4 & 62.9 \\
    		Sun \etal \cite{sun:compositional:2017} ICCV'17 $(+)$ & 52.8 & 54.8 & 54.2 & 54.3 & 61.8 & 67.2 & 53.1 & 53.6 & 71.7 & 86.7 & 61.5 & 53.4 & 61.6 & 47.1 & 53.4 & 59.1 \\
    		Fang \etal \cite{fang:learning:2018} AAAI'18 & 50.1 & 54.3 & 57.0 & 57.1 & 66.6 & 73.3 & 53.4 & 55.7 & 72.8 & 88.6 & 60.3 & 57.7 & 62.7 & 47.5 & 50.6 & 60.4 \\
    		Pavlakos \etal \cite{pavlakos:ordinal:2018} CVPR'18 $(+)$ & 48.5 & 54.4 & 54.4 & 52.0 & 59.4 & 65.3 & 49.9 & 52.9 & 65.8 & 71.1 & 56.6 & 52.9 & 60.9 & 44.7 & 47.8 & 56.2 \\
    		Yang \etal \cite{yang:3d:2018} CVPR'18 $(+)$ & 51.5 & 58.9 & 50.4 & 57.0 & 62.1 & 65.4 & 49.8 & 52.7 & 69.2 & 85.2 & 57.4 & 58.4 & 43.6 & 60.1 & 47.7 & 58.6 \\
    		Luvizon \etal \cite{luvizon:2d:2018} CVPR'18 $(\ast)(+)$ & 49.2 & 51.6 & 47.6 & 50.5 & 51.8 & 60.3 & 48.5 & 51.7 & 61.5 & 70.9 & 53.7 & 48.9 & 57.9 & 44.4 & 48.9 & 53.2 \\
    		Hossain \& Little \cite{hossain:exploiting:2018} ECCV'18 $(\dagger)(\ast)$ & 48.4 & 50.7 & 57.2 & 55.2 & 63.1 & 72.6 & 53.0 & 51.7 & 66.1 & 80.9 & 59.0 & 57.3 & 62.4 & 46.6 & 49.6 & 58.3 \\
    		Lee \etal \cite{lee:propagating:2018} ECCV'18 $(\dagger)(\ast)$ & \textbf{40.2} & 49.2 & 47.8 & 52.6 & 50.1 & 75.0 & 50.2 & \textbf{43.0} & \textbf{55.8} & 73.9 & 54.1 & 55.6 & 58.2 & 43.3 & 43.3 & 52.8 \\
    		\midrule
    		Ours, single-frame & 47.1 & 50.6 & 49.0 & 51.8 & 53.6 & 61.4 & 49.4 & 47.4 & 59.3 & 67.4 & 52.4 & 49.5 & 55.3 & 39.5 & 42.7 & 51.8 \\
    		Ours, 243 frames, causal conv. $(\dagger)$ & 45.9 & 48.5 & 44.3 & 47.8 & 51.9 & 57.8 & 46.2 & 45.6 & 59.9 & 68.5 & 50.6 & 46.4 & 51.0 & 34.5 & 35.4 & 49.0 \\
    		Ours, 243 frames, full conv. $(\dagger)$ & 45.2 & \textbf{46.7} & \underline{43.3} & \textbf{45.6} & \textbf{48.1} & \textbf{55.1} & \underline{44.6} & \underline{44.3} & 57.3 & \textbf{65.8} & \textbf{47.1} & \textbf{44.0} & \textbf{49.0} & \underline{32.8} & \textbf{33.9} & \textbf{46.8} \\
    		Ours, 243 frames, full conv. $(\dagger)(\ast)$ & \underline{45.1} & \underline{47.4} & \textbf{42.0} & \underline{46.0} & \underline{49.1} & \underline{56.7} & \textbf{44.5} & 44.4 & \underline{57.2} & \underline{66.1} & \underline{47.5} & \underline{44.8} & \underline{49.2} & \textbf{32.6} & \underline{34.0} & \underline{47.1}  \\
    		\bottomrule
    		\end{tabular}
    	}
    	\caption{Protocol 1: reconstruction error (MPJPE).}
    	\label{tbl:pose_results_best_p1}
    	\vspace{3mm}
	\end{subtable}
	\begin{subtable}{\linewidth}
	    \centering
    	\small
    	\tabcolsep=1mm
    	\resizebox{\textwidth}{!}{
    		\begin{tabular}{@{}l|rrrrrrrrrrrrrrr|r@{}}
    		& Dir. & Disc. & Eat & Greet & Phone & Photo & Pose & Purch. & Sit & SitD. & Smoke & Wait & WalkD. & Walk & WalkT. & Avg\\
    		\midrule
    		Martinez \etal \cite{martinez:simple:2017} ICCV'17 $(\ast)$ & 39.5 & 43.2 & 46.4 & 47.0 & 51.0 & 56.0 & 41.4 & 40.6 & 56.5 & 69.4 & 49.2 & 45.0 & 49.5 & 38.0 & 43.1 & 47.7 \\
    		Sun \etal \cite{sun:compositional:2017} ICCV'17 $(+)$ & 42.1 & 44.3 & 45.0 & 45.4 & 51.5 & 53.0 & 43.2 & 41.3 & 59.3 & 73.3 & 51.0 & 44.0 & 48.0 & 38.3 & 44.8 & 48.3 \\
    		Fang \etal \cite{fang:learning:2018} AAAI'18 & 38.2 & 41.7 & 43.7 & 44.9 & 48.5 & 55.3 & 40.2 & 38.2 & 54.5 & 64.4 & 47.2 & 44.3 & 47.3 & 36.7 & 41.7 & 45.7 \\
    		Pavlakos \etal \cite{pavlakos:ordinal:2018} CVPR'18 $(+)$ & 34.7 & 39.8 & 41.8 & 38.6 & 42.5 & 47.5 & 38.0 & 36.6 & 50.7 & 56.8 & 42.6 & 39.6 & 43.9 & 32.1 & 36.5 & 41.8 \\
    		Yang \etal \cite{yang:3d:2018} CVPR'18 $(+)$ & \textbf{26.9} & \textbf{30.9} & 36.3 & 39.9 & 43.9 & 47.4 & \textbf{28.8} & \textbf{29.4} & \textbf{36.9} & 58.4 & 41.5 & \textbf{30.5} & \textbf{29.5} & 42.5 & 32.2 & 37.7 \\
            Hossain \& Little \cite{hossain:exploiting:2018} ECCV'18 $(\dagger)(\ast)$ & 35.7 & 39.3 & 44.6 & 43.0 & 47.2 & 54.0 & 38.3 & 37.5 & 51.6 & 61.3 & 46.5 & 41.4 & 47.3 & 34.2 & 39.4 & 44.1 \\
    		\midrule
    		Ours, single-frame & 36.0 & 38.7 & 38.0 & 41.7 & 40.1 & 45.9 & 37.1 & 35.4 & 46.8 & 53.4 & 41.4 & 36.9 & 43.1 & 30.3 & 34.8 & 40.0 \\
    		Ours, 243 frames, causal conv. $(\dagger)$ & 35.1 & 37.7 & 36.1 & 38.8 & 38.5 & 44.7 & 35.4 & 34.7 & 46.7 & 53.9 & 39.6 & 35.4 & 39.4 & 27.3 & 28.6 & 38.1 \\
    		Ours, 243 frames, full conv. $(\dagger)$ & \underline{34.1} & \underline{36.1} & \underline{34.4} & \textbf{37.2} & \textbf{36.4} & \textbf{42.2} & \underline{34.4} & 33.6 & \underline{45.0} & \textbf{52.5} & \textbf{37.4} & \underline{33.8} & \underline{37.8} & \textbf{25.6} & \textbf{27.3} & \textbf{36.5} \\
    		Ours, 243 frames, full conv. $(\dagger)(\ast)$ & 34.2 & 36.8 & \textbf{33.9} & \underline{37.5} & \underline{37.1} & \underline{43.2} & \underline{34.4} & \underline{33.5} & 45.3 & \underline{52.7} & \underline{37.7} & 34.1 & 38.0 & \underline{25.8} & \underline{27.7} & \underline{36.8}  \\
    		\bottomrule
    		\end{tabular}
    	}
    	\caption{Protocol 2: reconstruction error after rigid alignment with the ground truth (P-MPJPE), where available.}
    	\label{tbl:pose_results_best_p2}
	\end{subtable}
	\caption{Reconstruction error on Human3.6M. \textbf{Legend:} $(\dagger)$ uses temporal information. $(\ast)$ ground-truth bounding boxes. $(+)$ extra data -- \cite{sun:compositional:2017, pavlakos:ordinal:2018, yang:3d:2018, luvizon:2d:2018} use 2D annotations from the MPII dataset, \cite{pavlakos:ordinal:2018} uses additional data from the Leeds Sports Pose (LSP) dataset as well as ordinal annotations. \cite{sun:compositional:2017, luvizon:2d:2018} evaluate every 64th frame. \cite{hossain:exploiting:2018} provided us with corrected results over the originally published results \protect\footnotemark. Lower is better, best in bold, second best underlined.}
    \label{tbl:pose_results_best}
\end{table*}

\section{Results}
\label{sec:pose_results}

\subsection{Temporal dilated convolutional model}

Table~\ref{tbl:pose_results_best} shows results for our convolutional model with $B=4$ blocks and a receptive field of 243 input frames for both evaluation protocols (\textsection\ref{sec:setup}).
The model has lower average error than all other approaches under both protocols, and does not rely on additional data such as many other approaches $(+)$.
Under protocol 1 (Table~\ref{tbl:pose_results_best_p1}), our model outperforms the previous best result \cite{lee:propagating:2018} by 6 mm on average, corresponding to an 11\% error reduction. 
Notably, \cite{lee:propagating:2018} uses ground-truth boxes whereas our model does not.

\begin{figure*}
	\centering
	\includegraphics[width=0.48\textwidth]{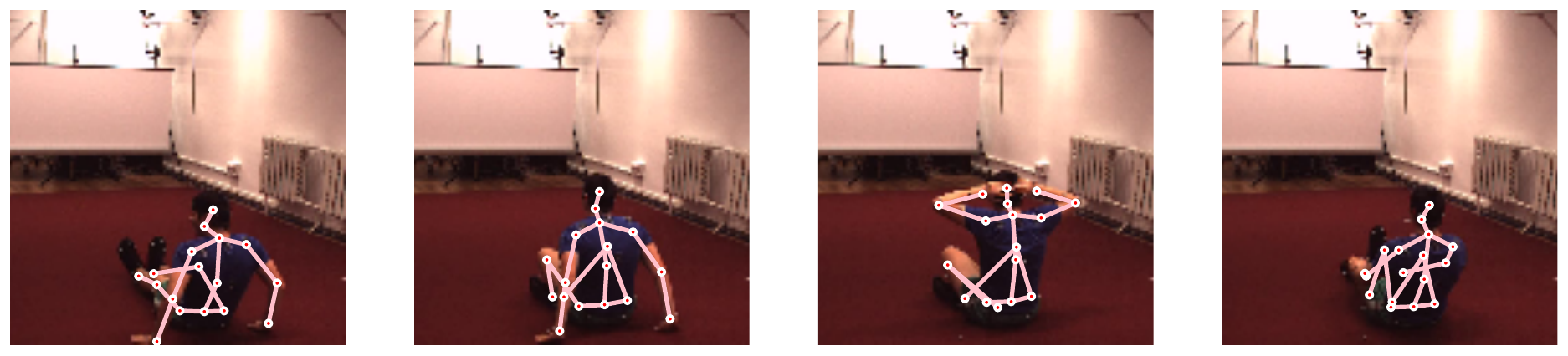}\quad
	\includegraphics[width=0.48\textwidth]{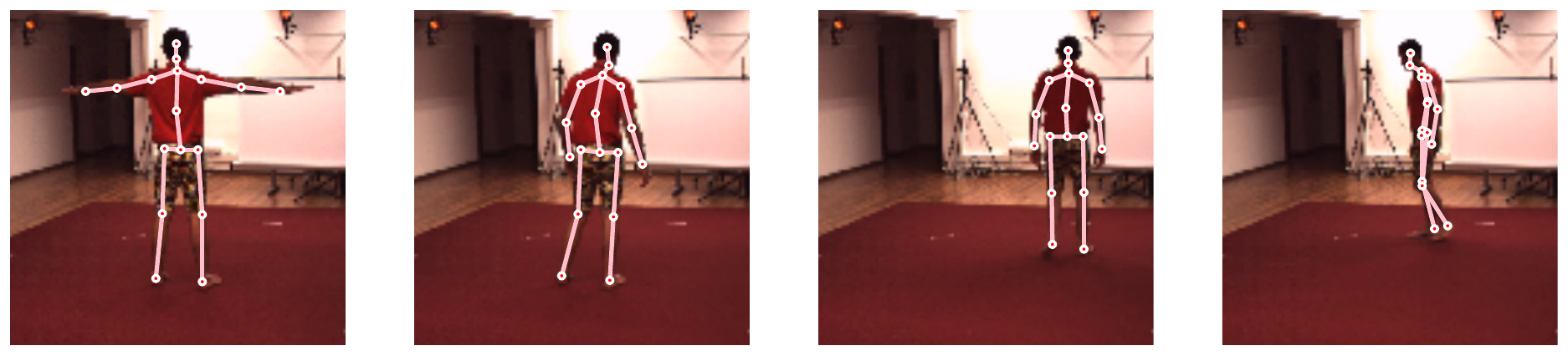}
	\includegraphics[width=0.48\textwidth]{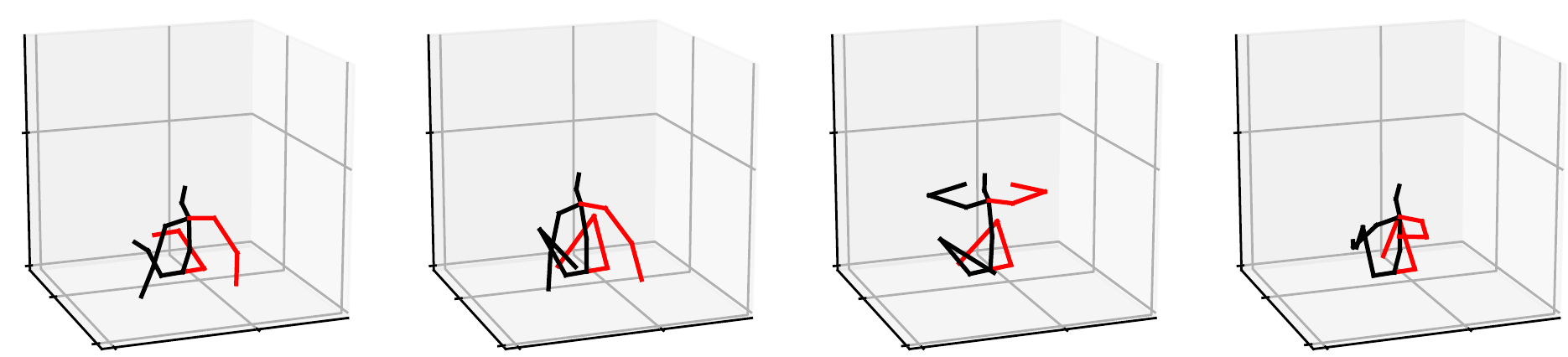}\quad
	\includegraphics[width=0.48\textwidth]{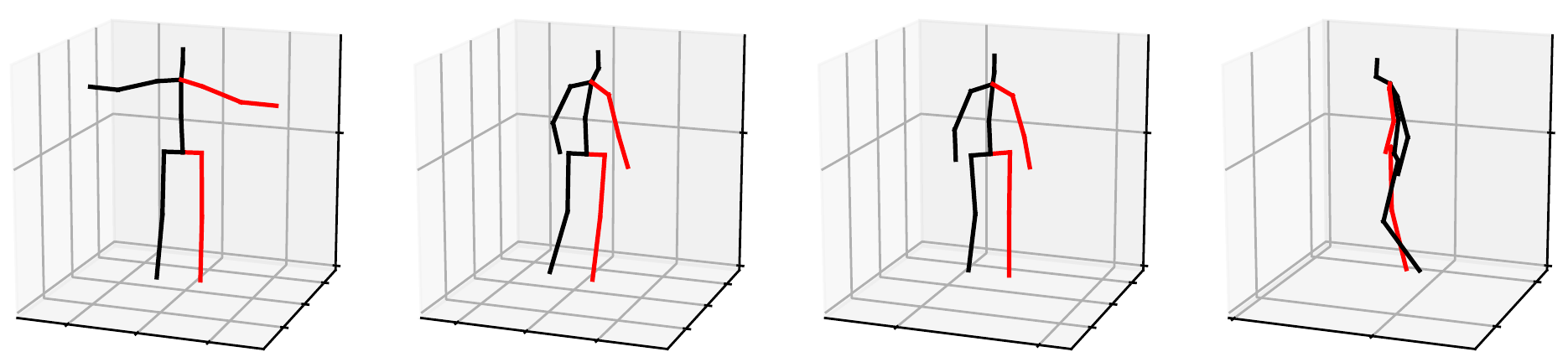}
    
	\caption{Qualitative results for two videos. \textbf{Top:} video frames with 2D pose overlay. \textbf{Bottom:} 3D reconstruction.}
	\label{fig:qualitative_pose}
\end{figure*}

The model clearly takes advantage of temporal information as the error is about 5 mm higher on average for protocol 1 compared to a single-frame baseline where we set the width of all convolution kernels to $W=1$. 
The gap is larger for highly dynamic actions, such as ``Walk'' (6.7 mm) and ``Walk Together'' (8.8 mm).
The performance for a model with causal convolutions is about half way between the single frame baseline and our model; causal convolutions enable online processing by predicting the 3D pose for the rightmost input frame. 
Interestingly, ground-truth bounding boxes result in similar performance to predicted bounding boxes with Mask R-CNN, which suggests that predictions are almost-perfect in our single-subject scenario.
Figure~\ref{fig:qualitative_pose} shows examples of predicted poses including the predicted 2D keypoints and we included a video illustration in the supplementary material (Appendix~\ref{app:demo}) as well as at \url{https://dariopavllo.github.io/VideoPose3D}.
\footnotetext{All subsequent results for \cite{hossain:exploiting:2018} in this paper were computed by us using their public implementation.}

\begin{table*}
	\centering
	\small
	\tabcolsep=1mm
	\resizebox{\textwidth}{!}{
		\begin{tabular}{@{}l|rrrrrrrrrrrrrrr|r@{}}
		& Dir. & Disc. & Eat & Greet & Phone & Photo & Pose & Purch. & Sit & SitD. & Smoke & Wait & WalkD. & Walk & WalkT. & Avg\\
		\midrule
		Single-frame & 12.8 & 12.6 & 10.3 & 14.2 & 10.2 & 11.3 & 11.8 & 11.3 & 8.2 & 10.2 & 10.3 & 11.3 & 13.1 & 13.4 & 12.9 & 11.6 \\
		Temporal & 3.0 & 3.1 & 2.2 & 3.4 & 2.3 & 2.7 & 2.7 & 3.1 & 2.1 & 2.9 & 2.3 & 2.4 & 3.7 & 3.1 & 2.8 & 2.8 \\
		\bottomrule
		\end{tabular}
	}
	\caption{Velocity error over the 3D poses generated by a convolutional model that considers time and a single-frame baseline.}
	\label{tbl:pose_results_velocity}
\end{table*}

Next, we evaluate the impact of the 2D keypoint detector on the final result.
Table~\ref{tbl:pose_results_martinez} reports accuracy of our model with ground-truth 2D poses, hourglass-network predictions from~\cite{martinez:simple:2017} (both pre-trained on MPII and fine-tuned on Human3.6M), Detectron and CPN (both pre-trained on COCO and fine-tuned on Human3.6M).
Both Mask R-CNN and CPN give better performance than the stacked hourglass network.
The improvement is likely to be due to the higher heatmap resolution, stronger feature combination (\emph{feature pyramid network} \cite{lin:feature:2017, ren:faster:2015} for Mask R-CNN and \emph{RefineNet} for CPN), and the more diverse dataset on which they are pretrained, \ie COCO \cite{lin:coco:2014}. 
When trained on 2D ground-truth poses, our model improves the lower bound of~\cite{martinez:simple:2017} by 8.3 mm, and the LSTM-based approach of Lee \etal \cite{lee:propagating:2018} by 1.2 mm for protocol 1.
Therefore, our improvements are not merely due to a better 2D detector.

\begin{table}
	\centering
	\small
	\tabcolsep=1mm
	\resizebox{\linewidth}{!}{
		\begin{tabular}{@{}l|r|r||l|r|r@{}}
		Method & P1 & P2 & Method & P1 & P2\\
		\midrule
		Martinez \etal \cite{martinez:simple:2017} (GT) & 45.5 & 37.1 & Ours (GT) & 37.2 & 27.2 \\
		Martinez \etal \cite{martinez:simple:2017} (SH PT) & 67.5 & 52.5 & Ours (SH PT from \cite{martinez:simple:2017}) & 58.6 & 45.0 \\
		Martinez \etal \cite{martinez:simple:2017} (SH FT) & 62.9 & 47.7 & Ours (SH FT from \cite{martinez:simple:2017}) & 53.4 & 40.1 \\
		Hossain \& Little \cite{hossain:exploiting:2018} (GT) & 41.6 & 31.7 & Ours (D PT) & 54.8 & 42.0 \\
        Lee \etal \cite{lee:propagating:2018} (GT) & 38.4 & -- & Ours (D FT) & 51.6 & 40.3 \\
		Ours (CPN PT) & 52.1 & 40.1 & Ours (CPN FT) & 46.8 & 36.5 \\
		\bottomrule
		\end{tabular}
	}
	\caption{Effect of the 2D detector on the final result, under Protocol 1 (P1) and Protocol 2 (P2) \textbf{Legend:} ground-truth (GT), stacked hourglass (SH), Detectron (D), cascaded pyramid network (CPN), pre-trained (PT), fine-tuned (FT).}
	\label{tbl:pose_results_martinez}
\end{table}

Absolute position errors do not measure the smoothness of predictions over time, which is important for video.
To evaluate this, we measure joint velocity errors (MPJVE), corresponding to the MPJPE of the first derivative of the 3D pose sequences.
Table~\ref{tbl:pose_results_velocity} shows that our temporal model reduces the MPJVE of the single-frame baseline by 76\% on average resulting in vastly smoother poses.

\begin{table}
	\centering
	\small
	\tabcolsep=1mm
	\resizebox{\linewidth}{!}{
		\begin{tabular}{@{}l|rrr|rrr|rrr@{}}
		& \multicolumn{3}{c}{Walk} & \multicolumn{3}{c}{Jog} & \multicolumn{3}{c}{Box} \\
		& S1 & S2 & S3 & S1 & S2 & S3 & S1 & S2 & S3 \\
		\midrule
		Pavlakos \etal \cite{pavlakos:coarse:2017} (MA) & 22.3 & 19.5 & \underline{29.7} & 28.9 & 21.9 & 23.8 & -- & -- & -- \\
		Martinez \etal \cite{martinez:simple:2017} (SA) & 19.7 & 17.4 & 46.8 & 26.9 & 18.2 & 18.6 & -- & -- & -- \\
		Pavlakos \etal \cite{pavlakos:ordinal:2018} (+) (MA) & 18.8 & 12.7 & \textbf{29.2} & 23.5 & 15.4 & 14.5  & -- & -- & -- \\
		Lee \etal \cite{lee:propagating:2018} (MA) & 18.6 & 19.9 & 30.5 & 25.7 & 16.8 & 17.7 & 42.8 & 48.1 & 53.4 \\
		\midrule
		Ours (SA) & \underline{14.5} & \underline{10.5} & 47.3  &  \underline{21.9} & \underline{13.4} & \underline{13.9}  &  \underline{24.3} & \underline{34.9} & \underline{32.1}  \\
		Ours (MA) & \textbf{13.9} & \textbf{10.2} & 46.6  &  \textbf{20.9} & \textbf{13.1} & \textbf{13.8}  &  \textbf{23.8} & \textbf{33.7} & \textbf{32.0} \\
		\bottomrule
		\end{tabular}
	}
	\caption{Error on HumanEva-I under Protocol 2 for single-action (SA) and multi-action (MA) models. Best in bold, second best underlined. (+) uses extra data. The high error on ``Walk'' of S3 is due to corrupted mocap data.}
	\label{tbl:humaneva_eval}
\end{table}

\begin{table}
	\centering
	\small
	\tabcolsep=1mm
	\resizebox{\linewidth}{!}{
		\begin{tabular}{@{}l|r|r|r@{}}
		Model & ~~~~~~~~~~~Parameters & ~~~~~~~~~~~$\approx$ FLOPs &~~~~~~~~~~~MPJPE \\
		\midrule
		Hossain \& Little \cite{hossain:exploiting:2018}~~~~~~~ & 16.96M & 33.88M & 41.6 \\
		\midrule
		Ours 27f w/o dilation & 29.53M & 59.03M & 41.1 \\ 
		\midrule
		Ours 27f & 8.56M & 17.09M & 40.6 \\ 
		Ours 81f & 12.75M & 25.48M & 38.7 \\
		Ours 243f & 16.95M & 33.87M & 37.8 \\
		\bottomrule
		\end{tabular}
	}
	\caption{Computational complexity of various models under \emph{Protocol 1} trained on ground-truth 2D poses. Results are without test-time augmentation.}
	\label{tbl:complexity}
\end{table}

Table~\ref{tbl:humaneva_eval} shows results on HumanEva-I and that our model generalizes to smaller datasets; results are based on pretrained Mask R-CNN 2D detections. 
Our models outperform the previous state-of-the-art.

Finally, Table~\ref{tbl:complexity} compares the convolutional model to the LSTM model of \cite{hossain:exploiting:2018} in terms of complexity. 
We report the number of model parameters and an estimate of the floating-point operations (FLOPs) to predict one frame at inference time (details in Appendix~\ref{app:complexity}). 
For the latter, we only consider matrix multiplications and report the amortized cost over a hypothetical sequence of infinite length (to disregard padding). 
MPJPE results are based on models trained on ground-truth 2D poses without test-time augmentation. 
Our model achieves a significantly lower error even when the number of computations are halved.
Our largest model with receptive field of 243 frames has roughly the same complexity as \cite{hossain:exploiting:2018}, but at 3.8 mm lower error. 
The table also highlights the effectiveness of dilated convolutions which increase complexity only logarithmically with respect to the receptive field.

Since our model is convolutional, it can be parallelized both over the number of sequences as well as over the temporal dimension. This contrasts to RNNs, which can only be parallelized over different sequences and are thus much less efficient for small batch sizes.
For inference, we measured about 150k FPS on a single NVIDIA GP100 GPU over a single long sequence, i.e., batch size one, assuming that 2D poses were already available. Speed is largely independent of the batch size due to parallel temporal processing.

\subsection{Semi-supervised approach}
\label{sec:semi_eval}

\begin{figure}
    \begin{subfigure}{\linewidth}
    	\centering
        \includegraphics[width=\linewidth]{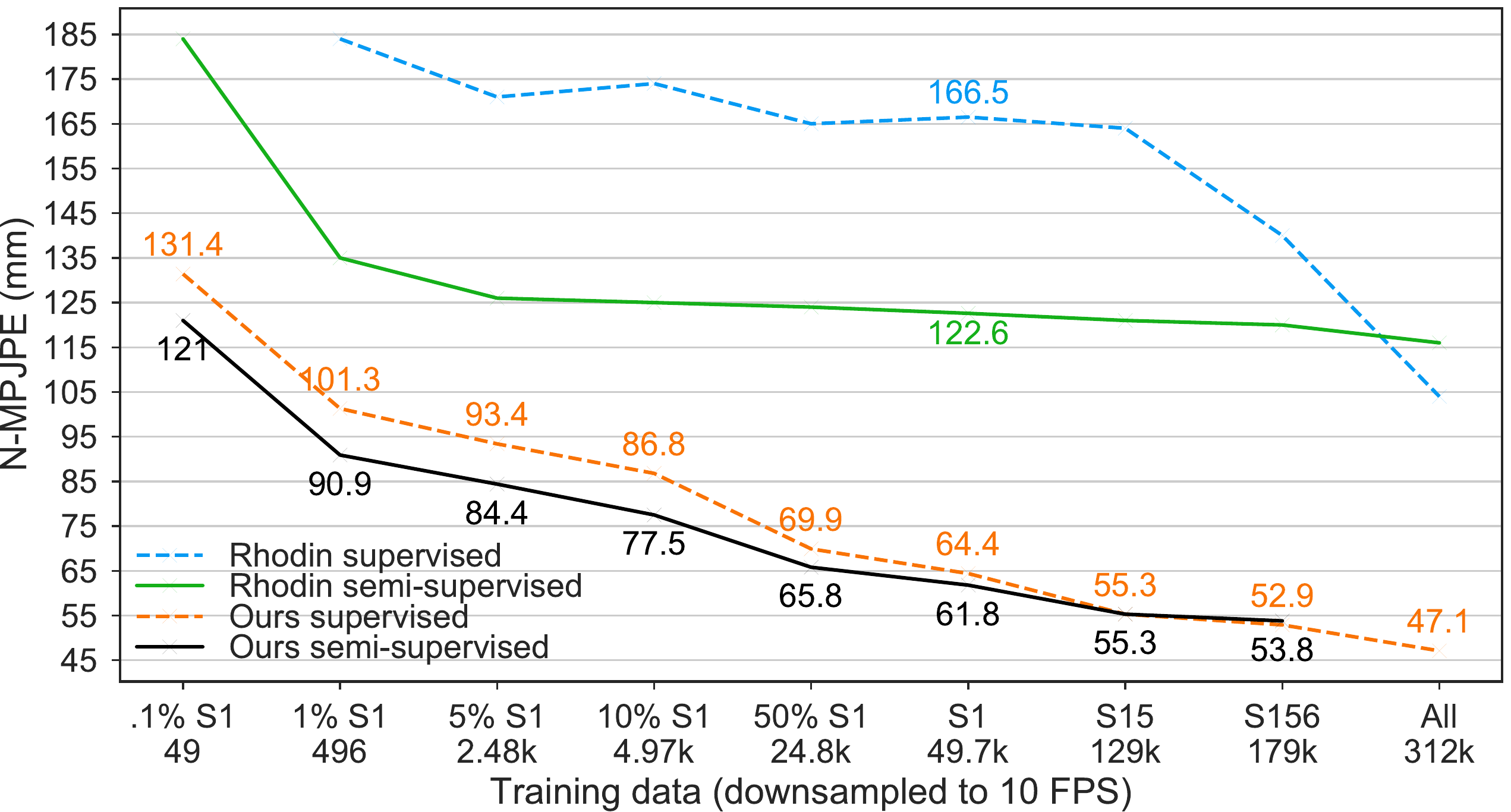}
        \vspace{-6mm}
        \caption{Downsampled to 10 FPS under Protocol 3.}
	    \label{fig:semi_downsampled}
    \end{subfigure}
    \vspace{2mm}
    
    \begin{subfigure}{\linewidth}
    	\centering
    	\includegraphics[width=\linewidth]{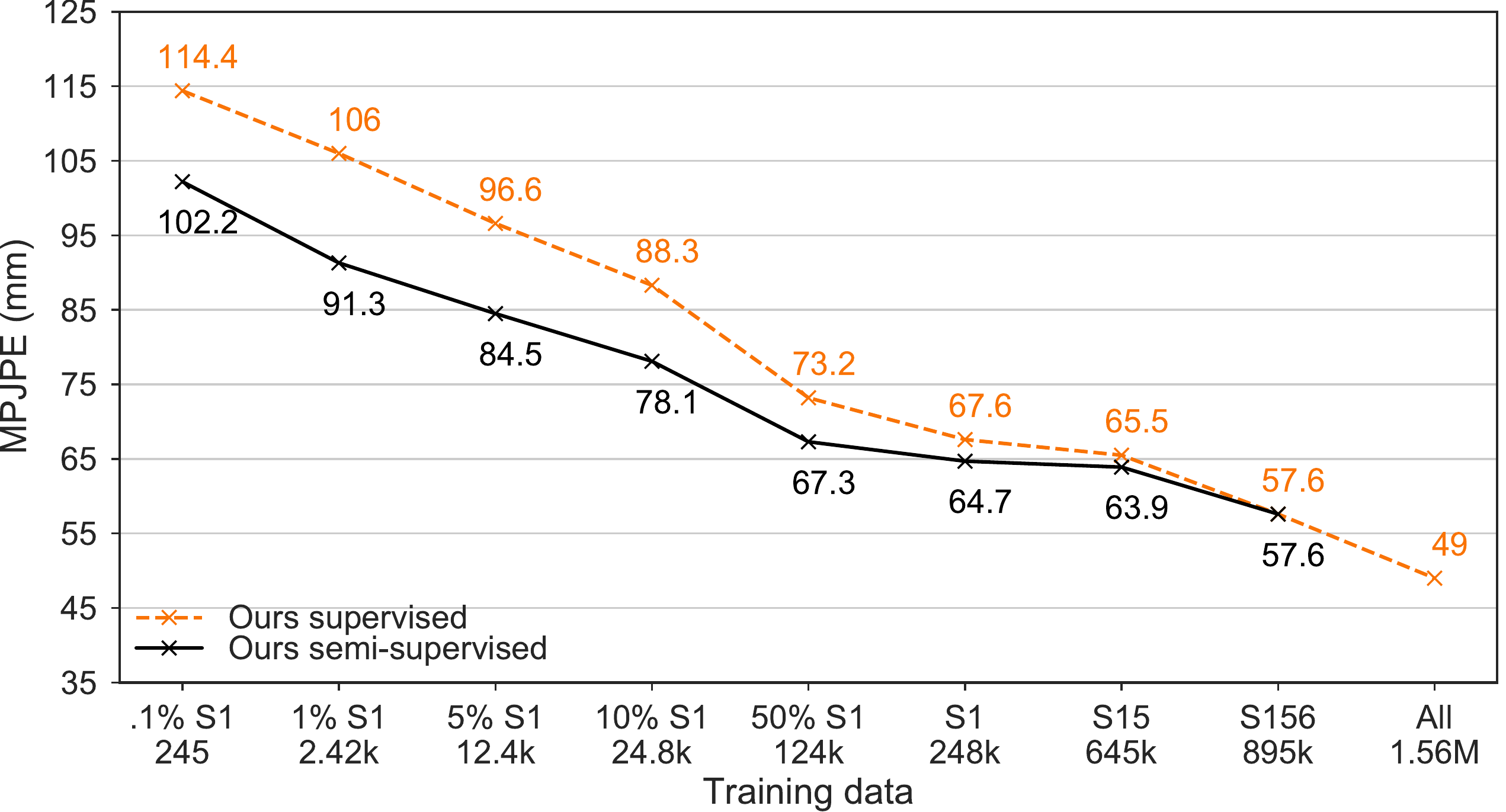}
    	\vspace{-6mm}
    	\caption{Full framerate under Protocol 1.}
    	\label{fig:semi_full}
    \end{subfigure}
	\vspace{2mm}
	
    \begin{subfigure}{\linewidth}
    	\centering
    	\includegraphics[width=\linewidth]{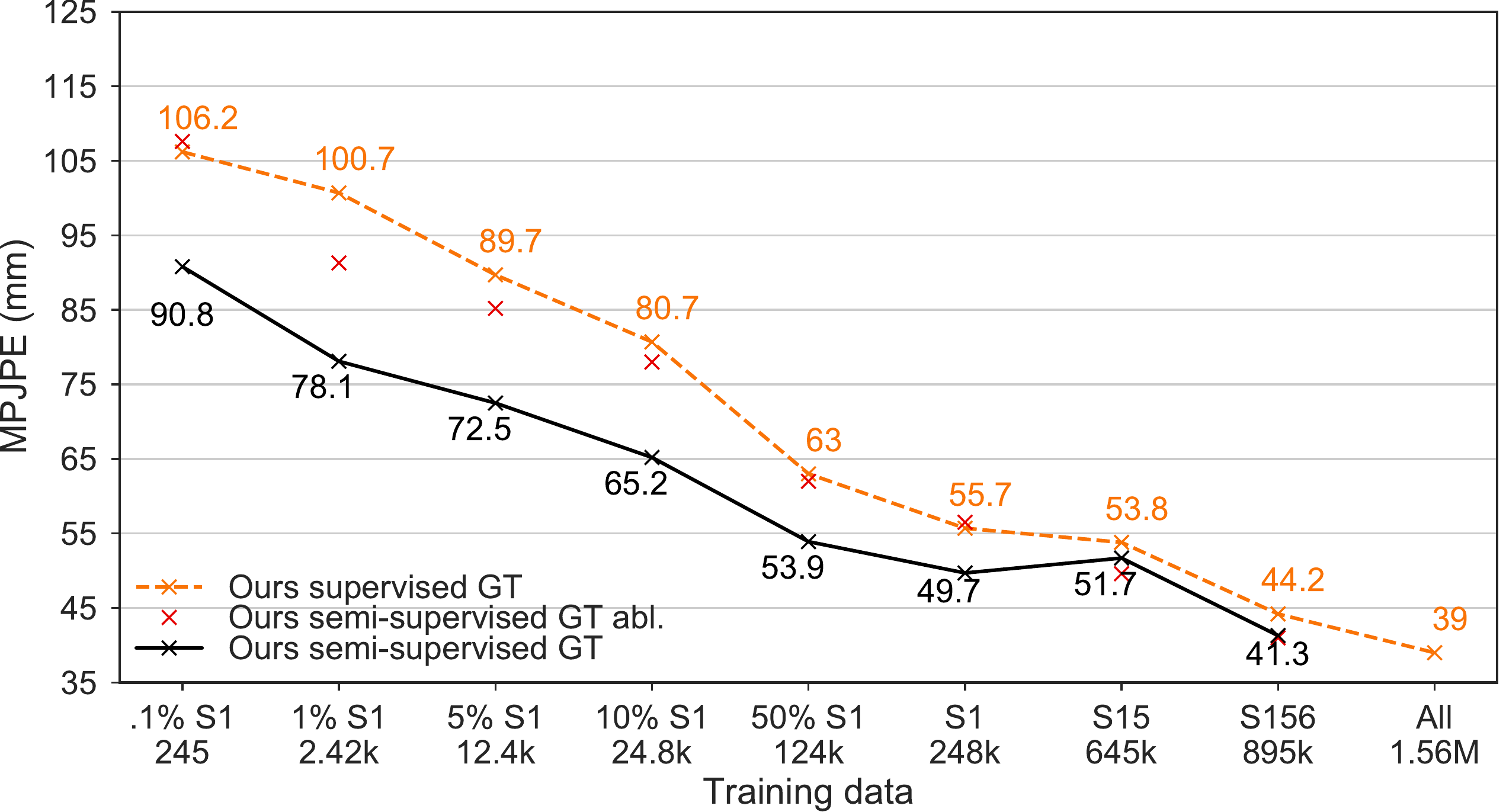}
    	\vspace{-6mm}
    	\caption{Full framerate under Protocol 1 with ground-truth 2D poses.}
    	\label{fig:semi_gt}
    \end{subfigure}
    	
	\caption{\textbf{Top:} comparison with \cite{rhodin:unsupervised:2018} on \emph{Protocol~3}, using a downsampled version of the dataset for consistency. 
	\textbf{Middle:} our method under \emph{Protocol~1} (full frame rate). 
	\textbf{Bottom:} our method under \emph{Protocol~1} when trained on ground-truth 2D poses (full frame rate). The small crosses (``abl.'' series) denote the ablation of the bone length term.}
	\label{fig:semi}
	\vspace{-3mm}
\end{figure}

We adopt the setup of \cite{rhodin:unsupervised:2018} who consider various subsets of the Human3.6M training set as labeled data and the remaining samples are used as unlabeled data.
Their setup also generally downsamples all data to 10 FPS (from 50 FPS).
Labeled subsets are created by first reducing the number of subjects and then by downsampling Subject 1.

Since the dataset is downsampled, we use a receptive field of 9 frames, equivalent to 45 frames upsampled. 
For the very small subsets, 1\% and 5\% of S1, we use 3 frames, and we use a single-frame model for 0.1\% of S1 where only 49 frames are available.
We fine-tuned CPN on the labeled data only and warm up training by iterating only over labeled data for a few epochs (1 epoch for $\geq$ S1, 20 epochs for smaller subsets).

Figure~\ref{fig:semi_downsampled} shows that our semi-supervised approach becomes more effective as the amount of labeled data decreases.
For settings with less than 5K labeled frames, our approach achieves improvements of about 9-10.4 mm N-MPJPE over our supervised baseline.
Our supervised baseline is much stronger than~\cite{rhodin:unsupervised:2018} and outperforms all of their results by a large margin. Although \cite{rhodin:unsupervised:2018} uses a single-frame model in all experiments, our findings still hold on 0.1\% of S1 (where we also use a single-frame model).

Figure~\ref{fig:semi_full} shows results for our method under the more common Protocol 1 for the non-downsampled version of the dataset (50 FPS). This setup is more appropriate for our approach since it allows us to exploit full temporal information in videos. Here we use a receptive field of 27 frames, except in 1\% of S1, where we use 9 frames, and 0.1\% of S1, where we use one frame. Our semi-supervised approach gains up to 14.7 mm MPJPE over the supervised baseline.

Figure~\ref{fig:semi_gt} switches the CPN 2D keypoints for ground-truth 2D poses to measure if we could perform better with a better 2D keypoint detector.
In this case, improvements can be up to 22.6 mm MPJPE (1\% of S1) which confirms that better 2D detections could improve performance.
The same graph shows that the bone length term is crucial for predicting valid poses, since it forces the model to respect kinematic constraints (line ``Ours semi-supervised GT abl.'').
Removing this term drastically decreases the effectiveness of semi-supervised training: for 1\% of S1 the error increases from 78.1 mm to 91.3 mm which compares to 100.7 mm for the supervised baseline.

\section{Conclusion}
\label{sec:ccl_pose}

We have introduced a simple fully convolutional model for 3D human pose estimation in video. Our architecture exploits temporal information with dilated convolutions over 2D keypoint trajectories. A second contribution of this work is back-projection, a semi-supervised training method to improve performance when labeled data is scarce. 
The method works with unlabeled video and only requires intrinsic camera parameters, making it practical in scenarios where motion capture is challenging (\eg outdoor sports).

Our fully convolutional architecture improves the previous best result on the popular Human3.6M dataset by 6mm average joint error which corresponds to a relative reduction of 11\% and also shows improvements on HumanEva-I.
Back-projection can improve 3D pose estimation accuracy by about 10mm N-MPJPE (15mm MPJPE) over a strong baseline when 5K or fewer annotated frames are available.

{\small
\bibliographystyle{ieee}
\bibliography{egbib}
}

\clearpage
\newpage
\appendix
\section{Supplementary material}
\subsection{Dilated convolutions and information flow}
\label{app:flow}

Dilated convolutions are a particular form of convolution with a sparse structure, whose kernel points are spaced uniformly and filled with zeros in between. For instance, the discrete filter $h = [\;1\; \fbox{2}\; 3\;]$ (where \fbox{2} is the center) becomes $[\;1\; 0\; \fbox{2}\; 0\; 3\;]$ with dilation factor $D = 2$, and $[\;1\; 0\; 0\; \fbox{2}\; 0\; 0\; 3\;]$ with $D = 3$. This particular structure enables optimized implementations that skip computations over zero points.

Consider a discrete convolution of two signals $f$ (of length $N$) and $h$ (zero-centered, of length $2M+1$), which can be computed as
\begin{equation}
    (f*h)[n] = \sum_{m=-M}^{M} f[n-m]\, h[m]
\end{equation}
Instead of spacing the kernel explicitly and applying a regular convolution, a dilated convolution can be computed as
\begin{equation}
    (f*h_D)[n] = \sum_{m=-M}^{M} f[D(n-m)]\, h[m]
\end{equation}
yielding roughly the same computational cost as regular convolutions for the same number of non-zero entries, while increasing the receptive field.

\begin{figure}[h]
    \centering
    \begin{subfigure}{\linewidth}
        \centering
        \includegraphics[width=\linewidth]{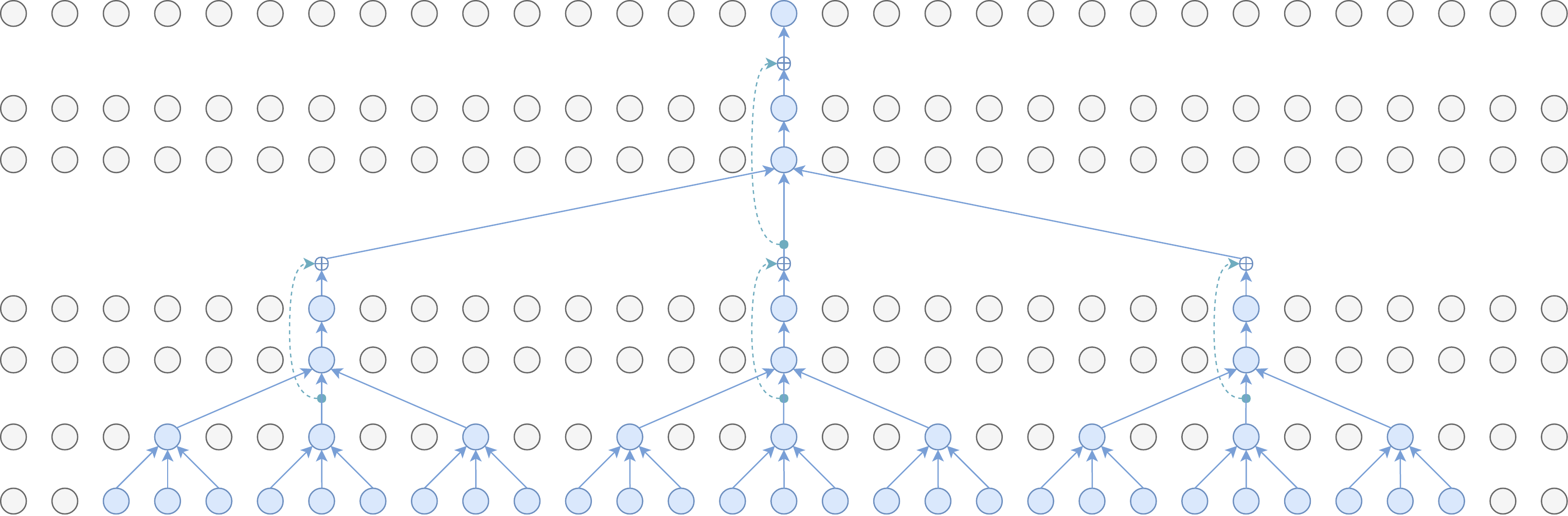}
        \caption{Symmetric convolutions.}
	    \label{fig:flow_normal}
    \end{subfigure}
    
    \vspace{3mm}
    
    \begin{subfigure}{\linewidth}
        \centering
        \includegraphics[width=\linewidth]{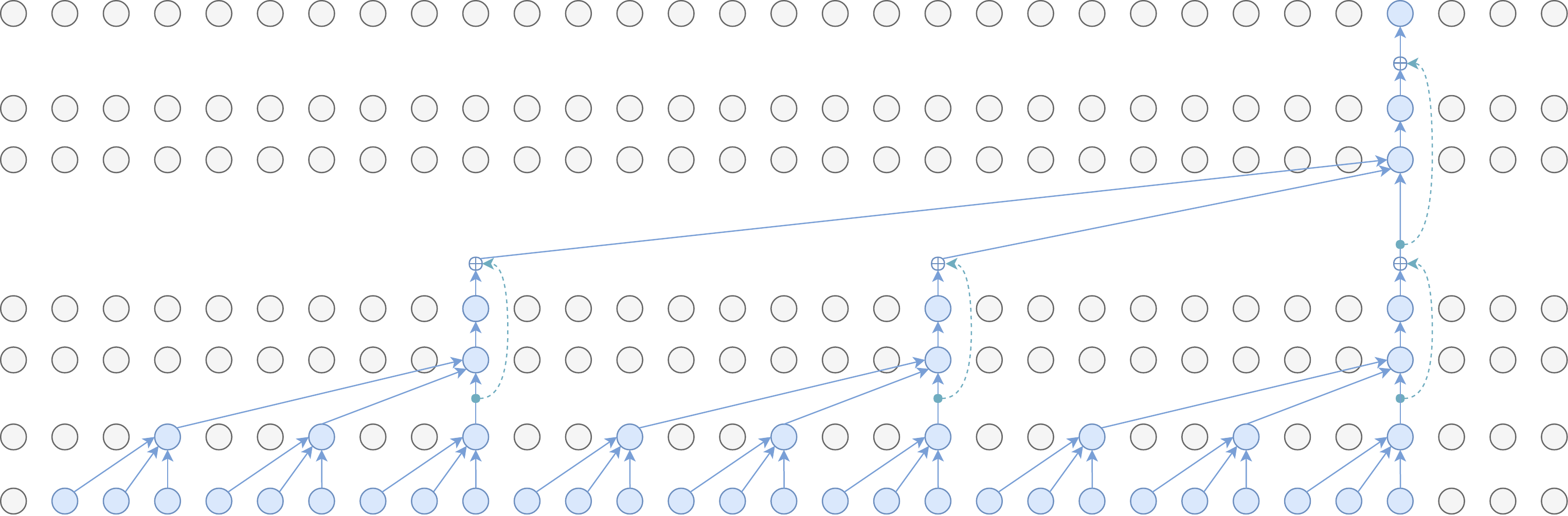}
        \caption{Causal convolutions.}
	    \label{fig:flow_causal}
    \end{subfigure}
	\caption{Information flow in our models, from input (bottom layer) to output (top layer). Dashed lines represent skip-connections.}
	\label{fig:flow}
\end{figure}

For illustration purposes, in \autoref{fig:flow} we depict the information flow in our models. We also highlight the difference between symmetric convolutions and causal convolutions.

\subsection{Computational cost analysis}
\label{app:complexity}

In this section we show how we computed the computational complexity of our model and that of \cite{hossain:exploiting:2018}. 
The common practice is to consider only matrix multiplications, as other operations (\eg biases, batch normalization, activations) have negligible impact on the final complexity. 
For~\cite{hossain:exploiting:2018}, we evaluated its reference implementation in TensorFlow. 
We computed the amortized cost to predict one frame using the TensorFlow profiler, and only counted operations corresponding to matrix multiplications. 
According to TensorFlow's approximation, multiplying a $N \times M$ matrix by a $M \times K$ matrix has a cost of $2\, N M K$ FLOPs (floating-point operations), which is equivalent to $N M K$ multiply-add operations.

For our model, we adopted the same convention. 
We provide a sample cost analysis for a model with a receptive field of 27 frames, which consists of 2 residual blocks. 
Since the matrix multiplications in our model are only due to convolutions, the analysis is straightforward and can be computed by hand.

\begin{figure}[h]
	\centering
	\includegraphics[width=\linewidth]{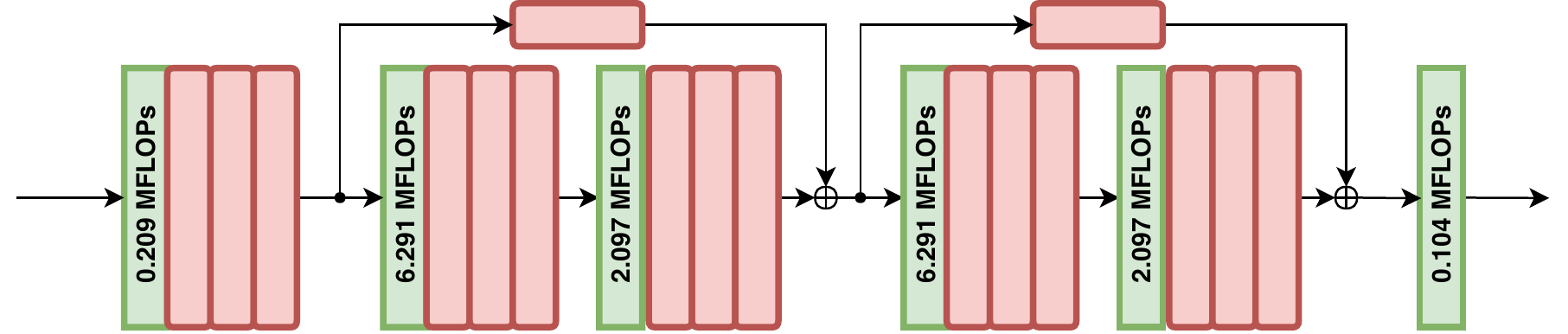}
	\caption{Architecture of a model with a receptive field of 27 frames, with the corresponding amortized cost to predict one frame in convolutional layers.}
	\label{fig:flops}
\end{figure}

As can be seen in \autoref{fig:flops}, the model consists of 6 convolutional layers. Disregarding padding (\ie for sequences of length $N \gg 0$), performing a 1D convolution with $C_{in}$ input channels, $C_{out}$ output channels, and width $W$ has a cost of $2\, N\, W\, C_{in}\, C_{out}$ FLOPs, \ie $2\, W\, C_{in}\, C_{out}$ FLOPs per frame. In our 27-frame model, the results can be summarized as follows:
\begin{enumerate}
    \item $W = 3$, channels $17 \cdot 2 \rightarrow 1024$, cost $0.209$ MFLOPs.
    \item $W = 3$, channels $1024 \rightarrow 1024$, cost $6.291$ MFLOPs.
    \item $W = 1$, channels $1024 \rightarrow 1024$, cost $2.097$ MFLOPs.
    \item $W = 3$, channels $1024 \rightarrow 1024$, cost $6.291$ MFLOPs.
    \item $W = 1$, channels $1024 \rightarrow 1024$, cost $2.097$ MFLOPs.
    \item $W = 1$, channels $1024 \rightarrow 17 \cdot 3$, cost $0.104$ MFLOPs.
\end{enumerate}
Total: $17.089$ MFLOPs per frame.

\subsection{Ablation of receptive field and channel size}
\label{sec:receptive_channel}

\begin{figure}
    \centering
        \begin{subfigure}{\linewidth}
        \centering
        \includegraphics[width=\linewidth]{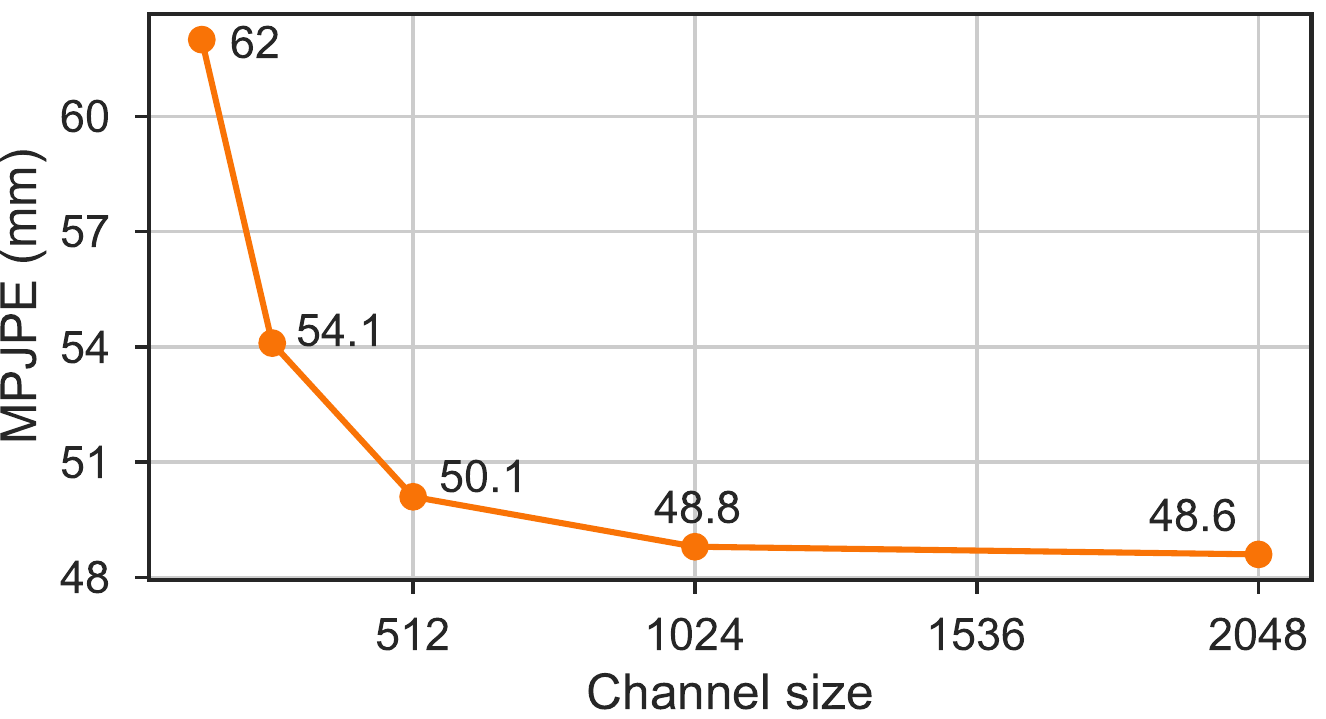}
        \caption{}
	    \label{fig:channel_size}
    \end{subfigure}\quad
    \begin{subfigure}{\linewidth}
        \centering
        \includegraphics[width=\linewidth]{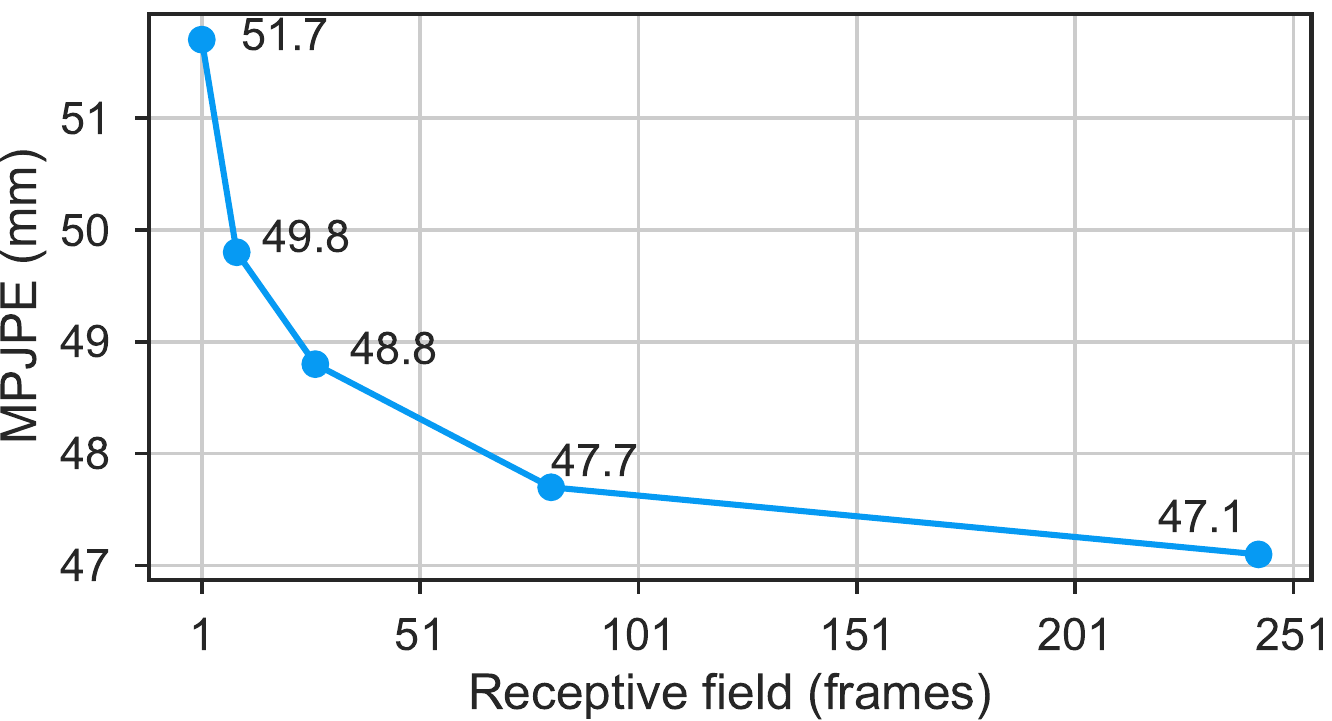}
        \caption{}
	    \label{fig:receptive_field}
    \end{subfigure}
	\caption{
		\textbf{Top:} Error as a function of the channel size, with a fixed receptive field of 27 frames.
		\textbf{Bottom:} Error as a function of the receptive field, with a fixed channel size of 1024. 
		Fine-tuned CPN detections for both experiments.
	}
\end{figure}

In \autoref{fig:receptive_field} we report the test error for different receptive fields, namely 1, 9, 27, 82, and 243 frames. 
To this end, we stack a varying number of residual blocks, each of which multiplies the receptive field by 3.
In the single-frame scenario, we use 2 blocks and set the convolution widths of all layers to 1, obtaining a model functionally equivalent to~\cite{martinez:simple:2017}. 
As can be seen, the model does not seem to overfit as the receptive field increases. 
On the other hand, the error tends to saturate quickly, suggesting that the task of 3D human pose estimation does not require modeling long-term dependencies. 
Therefore, we generally adopt a receptive field of 243 frames. 
Similarly, in \autoref{fig:channel_size} we vary the channel size between 128 and 2048, with the same findings: the model is not prone to overfitting, but the error saturates past a certain point. 
Since the computational complexity increases quadratically with respect to the channel size, we adopt $C = 1024$.

\subsection{Data augmentation and convolution type}
\label{app:data_conv}

When we remove test-time augmentation, the error increases to 47.7 mm (from 46.8 mm) in our top-performing model. 
If we also remove train-time augmentation, the error reaches 49.2 mm (another +1.5 mm).

Next, we replace dilated convolutions with regular dense convolutions. 
In a model with a receptive field of 27 frames and fine-tuned CPN detections, the error increases from 48.8 mm to 50.4 mm (+1.6 mm), while also increasing the number of parameters and computations by a factor of $\approx3.5$. 
This highlights that dilated convolutions are crucial for efficiency, and that they counteract overfitting.

\subsection{Batching strategy}
\label{app:batching}

We argue that the reconstruction error is strongly dependent on how the model is trained, and we suggested to generate minibatches in a way that only one output frame at a time is predicted. 
To show why this is important, we introduce a new hyperparameter -- the \emph{chunk size} $C$ (or \emph{step size}), which specifies how many frames are predicted at once per sample.
Predicting only one frame, i.e. $C = 1$, requires a full receptive field $F$ as input. 
Predicting two frames ($C = 2$) requires $F+1$ frames, and so on. 
It is evident that predicting multiple frames is computationally more efficient, as the results of intermediate convolutions can be shared among frames -- and in fact, we do this at inference. 
On the other hand, we show that during training this is detrimental to generalization. 

\begin{figure}
	\centering
    \begin{subfigure}{0.5\linewidth}
	    \centering
	    \includegraphics[width=\linewidth]{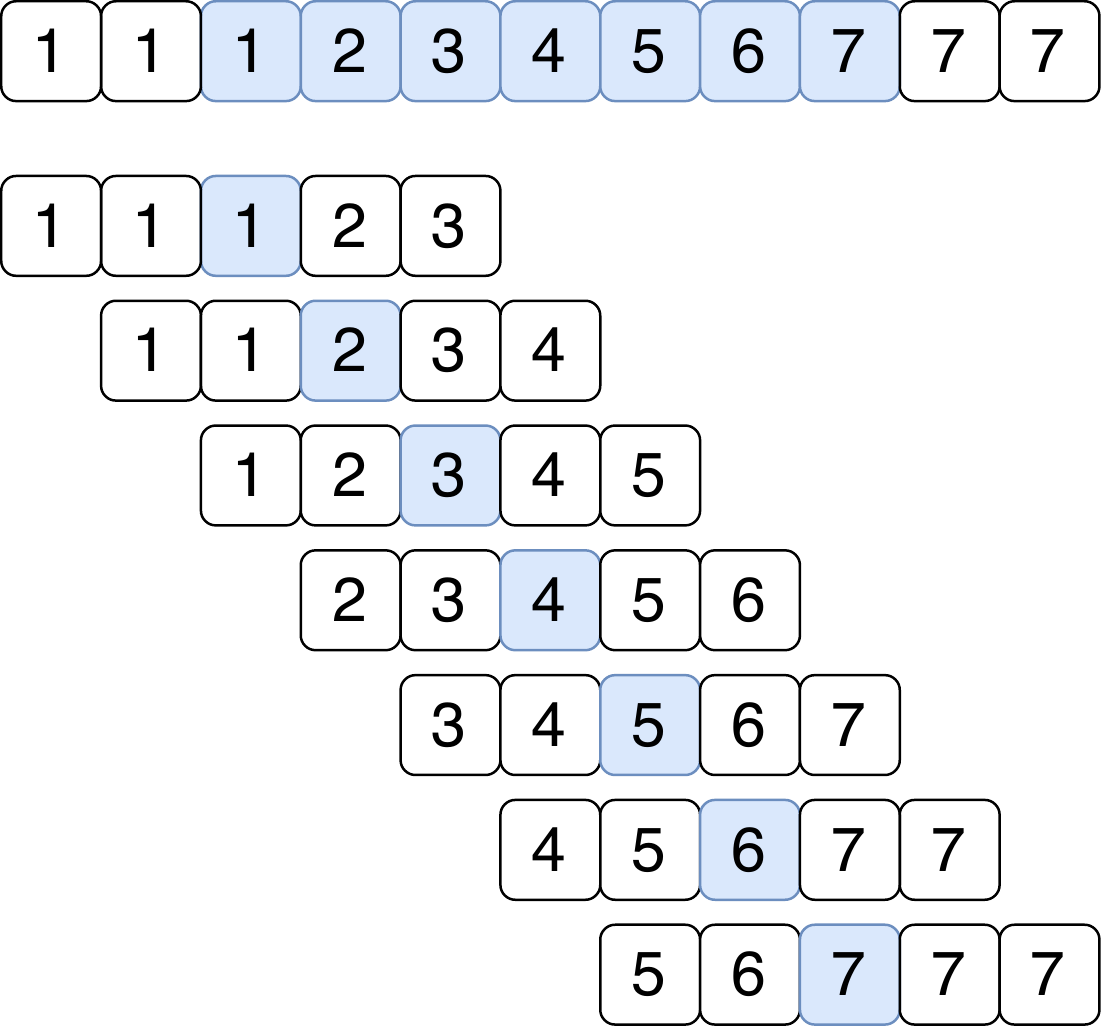}
    	\caption{}
        \label{fig:batching}
	\end{subfigure}
    \begin{subfigure}{\linewidth}
	    \centering
	    \includegraphics[width=\linewidth]{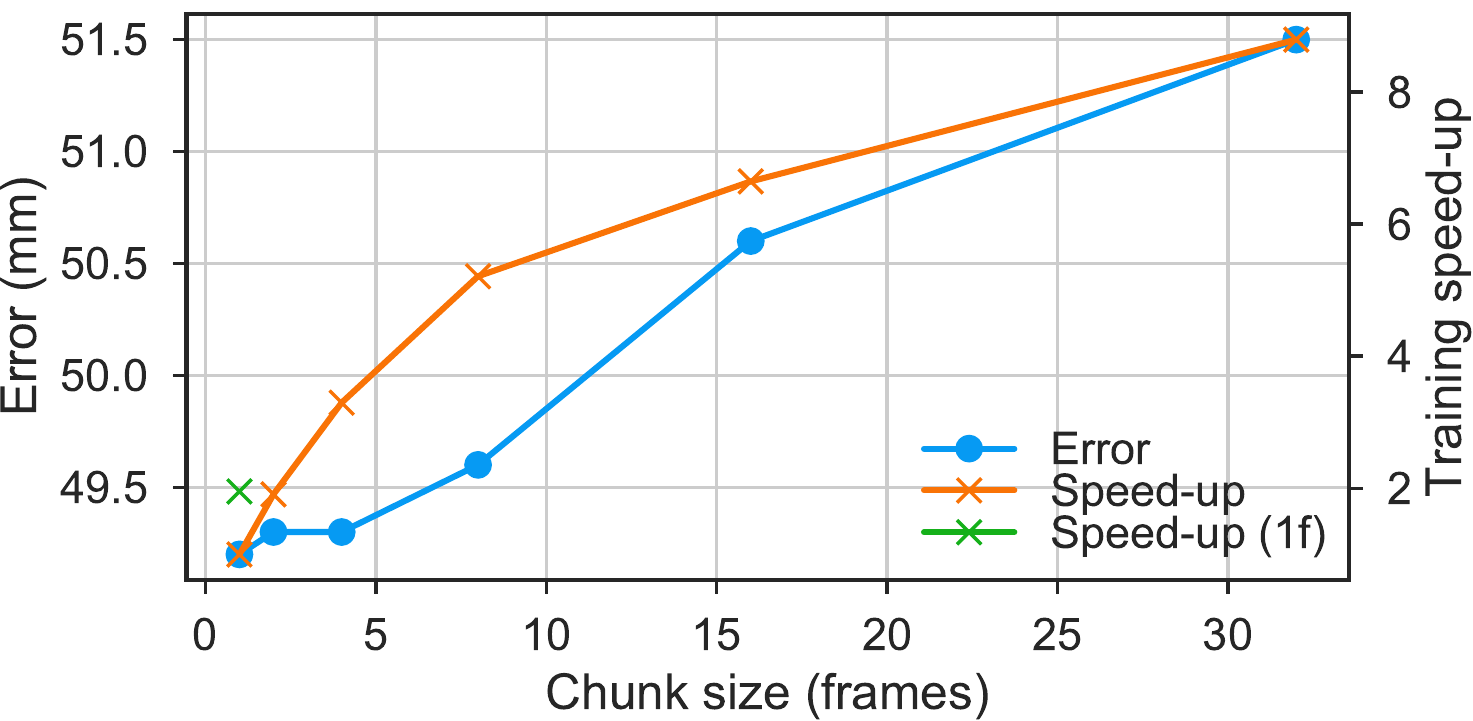}
		\caption{}
		\label{fig:step_size}
	\end{subfigure}
	\caption{\textbf{Top:} batch creation process \emph{for training}. 
	This example shows a video of 7 frames which is used to train a model with a receptive field of 5 frames. 
	We generate a training example for each of the 7 frames, such that only the center frame is predicted. 
	The video is padded by replicating boundary frames. 
	\textbf{Bottom:} reconstruction error and training speed-up with different step sizes. The speed-up is relative to $C = 1$. 
	The \texttt{1f} variant shows the speed up corresponding to the implementation optimized for single-frame predictions.
	}
\end{figure}

\autoref{fig:step_size} illustrates the reconstruction error (as well as the relative speedup in training time) when training a 27-frame model with different step sizes, namely 1, 2, 4, 8, 16, and 32 frames. 
Since predicting multiple frames is equivalent to increasing the batch size -- thus hurting generalization \cite{keskar:large:2017} -- we make the results comparable by adjusting the batch size so that the model always predicts 1024 frames. 
Therefore, the 1-frame experiment adopts a batch size of 1024 sequences, which becomes 512 for 2 frames, 256 for 4 frames, and so on. 
This methodology also ensures the models will be trained with the same number of weight updates.

The results show that the error decreases in conjunction with the step size, at the expense of training speed. 
The impaired performance of the models trained with high step size is caused by correlated batch statistics \cite{hoffer:norm:2018}. 
Our implementation optimized for single-frame outputs achieves a speed-up factor of $\approx 2$, but this gain is even higher on models with larger receptive fields (\eg $\approx 4$ with 81 frames), and enabled us to train the model with 243 frames.

\subsection{Optimized training implementation}
\label{app:optimize}

\begin{figure}[t!]
    \centering
        \begin{subfigure}{\linewidth}
        \centering
        \includegraphics[width=\linewidth]{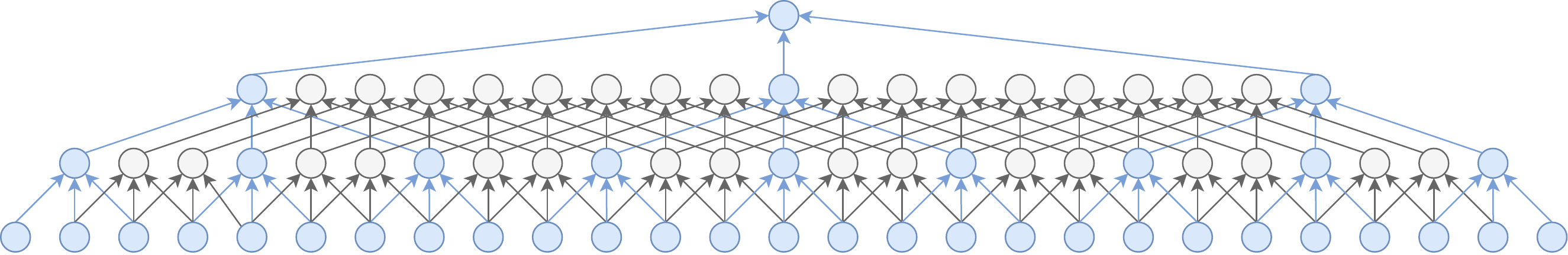}
        \caption{Layer-by-layer implementation.}
	    \label{fig:conv_wasted}
    \end{subfigure}
    
    \vspace{3mm}
    
    \begin{subfigure}{\linewidth}
        \centering
        \includegraphics[width=\linewidth]{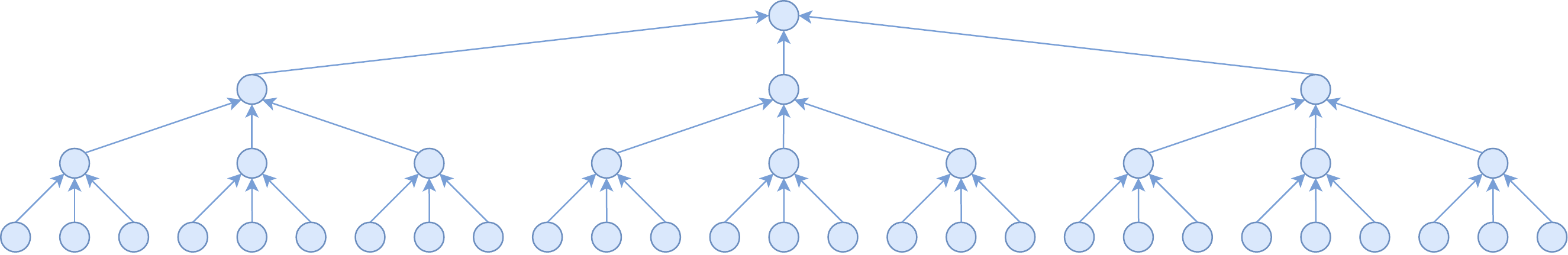}
        \caption{Implementation optimized for single-frame predictions.}
	    \label{fig:conv_optimized}
    \end{subfigure}
	\caption{Comparison between two implementations for a single-frame prediction, receptive field of 27 frames. In the layer-by-layer implementation many intermediate states are wasted, whereas the optimized implementation computes only required states. As the length of the sequence increases, the layer-by-layer implementation becomes more efficient.}
	\label{fig:optimized}
\end{figure}

\autoref{fig:optimized} shows why our implementation tailored for single-frame predictions is important. A regular implementation computes intermediate states layer by layer. This is very efficient for long sequences, as the states computed in layer $n$ can be reused by layer $n+1$ without recomputing them. However, for short sequences, this approach becomes inefficient because states near boundaries are not used. In the extreme case of single-frame predictions (which we use for training), many intermediate computations are wasted, as can be seen in \autoref{fig:conv_wasted}. In this case, we replace dilated convolutions with strided convolutions, making sure to obtain a model which is functionally equivalent to the original one (\eg by also adapting skip-connections). This strategy ensures that no intermediate states will be discarded.

As mentioned, at inference we use the regular layer-by-layer implementation since it is more efficient for multi-frame predictions.

\subsection{Demo videos}
\label{app:demo}

The supplementary material contains several videos highlighting the smoothness of the predictions of our temporal convolutional model compared to the single frame baseline. 
Specifically, we show side by side the original video sequence, poses predicted by the single-frame baseline, poses from the temporal convolutional model as well as the ground-truth poses. Some demo videos can also be found at \url{https://dariopavllo.github.io/VideoPose3D}.

\makeatletter
\setlength{\@fptop}{0pt}
\makeatother

\end{document}